\documentclass[10pt,twocolumn,letterpaper]{article}

\usepackage[pagenumbers]{cvpr} 

\definecolor{cvprblue}{rgb}{0.21,0.49,0.74}
\usepackage[pagebackref,breaklinks,colorlinks,allcolors=cvprblue]{hyperref}
\usepackage{multicol}
\usepackage{multirow}
\usepackage{adjustbox}
\usepackage[table]{xcolor}
\usepackage{xcolor}
\usepackage{tcolorbox}
\usepackage{makecell}
\usepackage{float}
\usepackage{soul}
\usepackage{placeins}
\usepackage{pifont}
\newcommand{\xmark}{\ding{55}}
\newcommand{\cmark}{\ding{51}}

\newcommand{\best}[1]{\textbf{#1}}
\newcommand{\second}[1]{\underline{#1}}

\usepackage{graphicx}
\usepackage{tabularx}
\usepackage{array}
\usepackage{setspace}
\usepackage{booktabs}

\newcolumntype{Y}{>{\raggedright\arraybackslash}X}

\usepackage[table]{xcolor}
\usepackage{booktabs}
\usepackage{makecell}
\usepackage[normalem]{ulem}
\usepackage{dashrule}

\definecolor{grid}{HTML}{D9DEE6}
\definecolor{tbhead}{HTML}{EFF3F9}
\definecolor{tbshade}{HTML}{F6F8FB}

\definecolor{critA}{HTML}{FFF3E0} 
\definecolor{critB}{HTML}{E3F2FD} 
\definecolor{critC}{HTML}{E8F5E9} 
\definecolor{critD}{HTML}{ECE7F6} 
\definecolor{critE}{HTML}{FCE4EC} 
\definecolor{critF}{HTML}{FFFDE7} 

\definecolor{critTextA}{HTML}{D97E00} 
\definecolor{critTextB}{HTML}{1565C0} 
\definecolor{critTextC}{HTML}{2E7D32} 
\definecolor{critTextD}{HTML}{512DA8} 
\definecolor{critTextE}{HTML}{C2185B} 
\definecolor{critTextF}{HTML}{F9A825} 

\definecolor{Acol}{HTML}{E8F5E9}
\definecolor{Atext}{HTML}{2E7D32}
\definecolor{Bcol}{HTML}{E3F2FD}
\definecolor{Btext}{HTML}{1565C0}

\usepackage{array,ragged2e} %
\usepackage{booktabs}
\usepackage[dvipsnames]{xcolor}
\usepackage{colortbl}
\definecolor{panelyellow}{RGB}{255,252,230}
\definecolor{panelgreen}{RGB}{235,250,235}
\definecolor{best}{RGB}{255,220,180}       
\definecolor{second}{RGB}{255,244,224}     
\definecolor{ForestGreen}{RGB}{34,139,34}
\definecolor{BrickRed}{RGB}{178,34,34}

\def\paperID{} 
\def\confName{CVPR}
\def\confYear{2026}

\title{Multi-Crit: Benchmarking Multimodal Judges on Pluralistic Criteria-Following}

\author{Tianyi Xiong\textsuperscript{1*},
Yi Ge\textsuperscript{1*}, 
Ming Li\textsuperscript{1}, 
Zuolong Zhang\textsuperscript{1}, 
Pranav Kulkarni\textsuperscript{1}, 
Kaishen Wang\textsuperscript{1}, 
Qi He\textsuperscript{1}, \\
Zeying Zhu\textsuperscript{1}, 
Chenxi Liu\textsuperscript{1}, 
Ruibo Chen\textsuperscript{1}, 
Tong Zheng\textsuperscript{1}, 
Yanshuo Chen\textsuperscript{1}, 
Xiyao Wang\textsuperscript{1}, \\
 Renrui Zhang, 
Wenhu Chen\textsuperscript{2}, 
Heng Huang\textsuperscript{1}
\vspace{1.3mm}\\
\textsuperscript{1}University of Maryland, College Park,~~\textsuperscript{2}University of Waterloo   \\
\href{https://multi-crit.github.io}{\texttt{multi-crit.github.io}} \\
}

\newcommand{\ours}{Multi-Crit\xspace}

\begin{document}

\let\oldtwocolumn\twocolumn
\renewcommand\twocolumn[1][]{%
    \oldtwocolumn[{#1}{
    \begin{center}
        \vspace{-6.4mm}
        \includegraphics[width=0.987\textwidth]{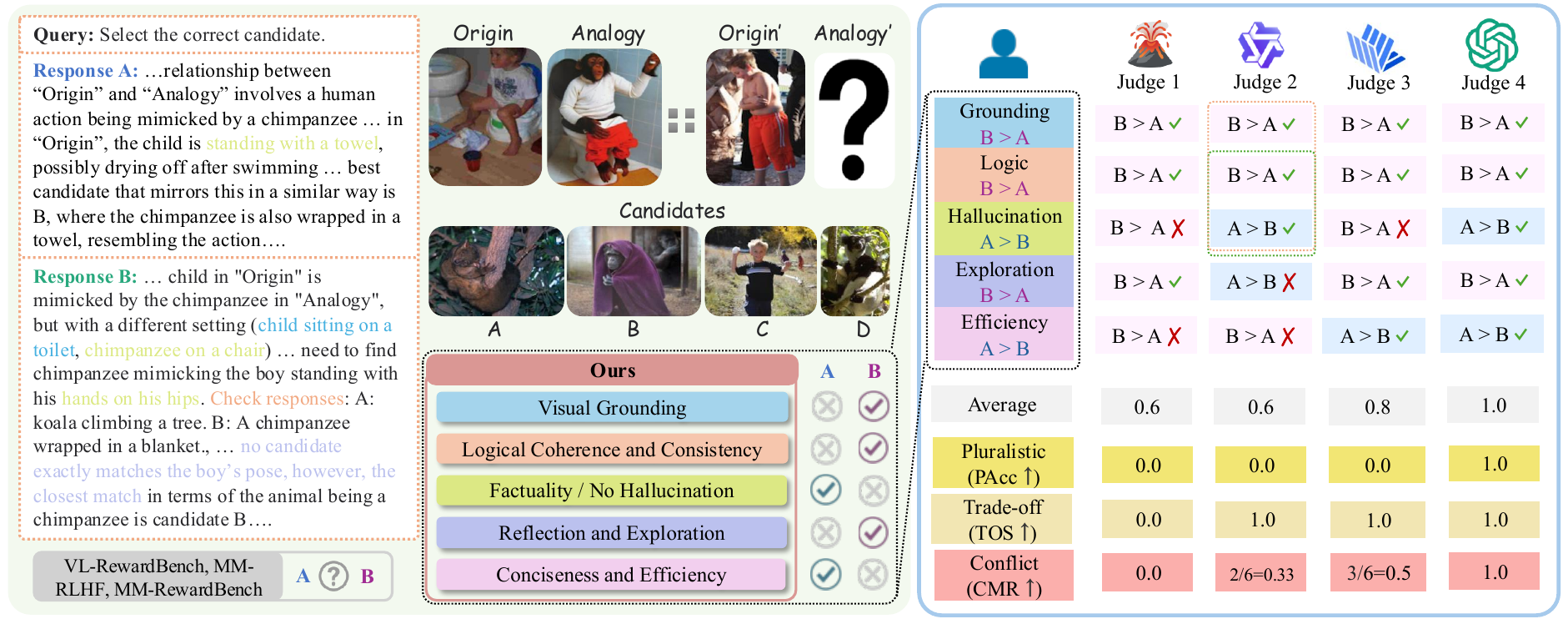}
        \vspace{-2.4mm}
        \captionof{figure}{
          \textbf{Overview of \ours.} \textit{Left}: Unlike prior works that assign a single overall preference label, \ours\ provides pluralistic, multi-criterion human judgments, exposing conflicts between different evaluation criteria within the same sample (e.g., Logic vs.\ No Hallucination, Reflection vs.\ Efficiency). \textit{Right}: We introduce three complementary metrics to systematically assess LMM judges on their ability to follow pluralistic evaluation criteria, recognize preference trade-offs, and capture criterion-level conflicts.
        }
        \label{fig:data_example}
        \vspace{0.2mm}
    \end{center}
    }]
}

\maketitle

\begin{NoHyper}
\renewcommand\thefootnote{*}
\footnotetext{Core contributors.}
\end{NoHyper}

\let\twocolumn\oldtwocolumn

\begin{abstract}
Large multimodal models (LMMs) are increasingly adopted as judges in multimodal evaluation systems due to their strong instruction following and consistency with human preferences. However, their ability to follow diverse, fine-grained evaluation criteria remains underexplored.
We develop \ours, a benchmark for evaluating multimodal judges on their capacity to follow pluralistic criteria and produce reliable criterion-level judgments. Covering both open-ended generation and verifiable reasoning tasks, \ours\ is built through a rigorous data curation pipeline that gathers challenging response pairs with multi-criterion human annotations. It further introduces three novel metrics for systematically assessing pluralistic adherence, criterion-switching flexibility, and the ability to recognize criterion-level preference conflicts.
Comprehensive analysis of 25 LMMs reveals that 1) proprietary models still struggle to maintain consistent adherence to pluralistic criteria--especially in open-ended evaluation; 2) open-source models lag further behind in flexibly following diverse criteria; and 3) critic fine-tuning with holistic judgment signals enhances visual grounding but fails to generalize to pluralistic criterion-level judgment.
Additional analyses on reasoning fine-tuning, test-time scaling, and boundary consistency between open-source and proprietary models further probe the limits of current multimodal judges.
As a pioneering study, \ours\ lays the foundation for building reliable and steerable multimodal AI evaluation.

\end{abstract}
    
\section{Introduction}
\label{sec:intro}

The recent advancements of large multimodal models (LMMs)~\cite{li2024llava,liu2023visual,bai2025qwen2,chen2024internvl,chen2025eagle,hurst2024gpt} in visual understanding~\cite{tong2024cambrian,yu2016modeling,yu2023mm,guan2024hallusionbench,wang2025damo,chen2023gpt} and reasoning~\cite{meng2025mm,zhang2024mathverse,jiang2025mme} have underscored the growing need for scalable and reliable multimodal evaluation systems.
However, evaluating free-form outputs in open-ended or reasoning-intensive tasks remains inherently challenging, as static or reference-based metrics often fail to generalize and provide limited insight into model behavior.
A growing trend has emerged to deploy LMMs themselves as generative judges (\textit{LMM-as-a-Judge})~\cite{zhang2023gpt, wang2025enhancing, xiong2025llava}. 
Given a multimodal prompt consisting of an image and a user query, model responses, and a set of predefined evaluation criteria, the judge model produces either scalar scores for individual responses or pairwise preference judgments, accompanied by textual justifications for its evaluation.
Owing to its scalability and flexibility in defining task-specific criteria, this paradigm has been widely adopted for open-ended evaluation across diverse multimodal benchmarks~\cite{lu2024wildvision, guan2024hallusionbench, zhang2024mathverse, ye2025painting, yu2025rlaif, li2024llava, jiang2025mme, sun2023aligning, xie2025vlms}. A series of works fine-tune open-source models to develop dedicated judge and critic capacities, and have proven effective in providing AI feedback for aligning LMMs through reinforcement fine-tuning~\cite{li2024vlfeedback, wang2025unified, liu2025improving, zheng2025parallel} or test-time scaling~\cite{zhang2025r1, wang2025llava, zheng2026parallel}.

The reliability of the \textit{LMM-as-a-Judge} paradigm relies on two key components:
(1) \textbf{consistency with human judgment}, and
(2) \textbf{flexibility to follow diverse, task-specific evaluation criteria} that capture different aspects of model behavior (e.g., factuality, visual grounding, structural coherence).
However, existing multimodal reward and judge benchmarks~\cite{li2025vl,zhang2025mm,chen2024mllm,zeng-etal-2025-mm} mainly address the first aspect by providing a single overall preference label for each pair of model responses.
This coarse formulation overlooks the evaluation of judge models in producing criterion-specific judgment. Specifically, two responses may exhibit trade-offs across different criteria—for instance, a concise but factually correct answer versus a visually grounded yet slightly hallucinated one.
This motivates a key question: {\em Can LMM-based judges follow diverse evaluation criteria and recognize conflicts among criterion-level judgments?}

To address this limitation, we introduce \ours, a challenging benchmark designed to comprehensively evaluate multimodal judges on their ability to follow diverse evaluation criteria and produce reliable, criterion-level judgments aligned with human evaluators.
As shown in Figure~\ref{fig:data_example}, each evaluation sample is annotated with multiple criterion-level human preferences, capturing distinct aspects of response quality and the potential conflicts between criteria.
\ours\ is built upon two key research questions:

\begin{enumerate}[itemsep=3pt,topsep=4pt,parsep=0pt,partopsep=0pt,leftmargin=10pt]
    \item \textbf{How to construct evaluation data with multi-criterion human preferences that capture diverse aspects of response quality and reveal intra-sample conflicts?}
    We design a rigorous data curation and preference-annotation pipeline that:  
    (1) collects diverse multimodal prompts spanning both open-ended generation and verifiable reasoning tasks;  
    (2) selects challenging response pairs from various strong LMMs with fine-grained criterion-level quality differences via multi-step filtering;  
    (3) defines clear evaluation criteria covering complementary aspects of multimodal judgment; and  
    (4) gathers high-quality human annotations ensuring reliability and inter-annotator consistency.
    
    \item \textbf{How to assess the performance of LMM-based judges in pluralistic criteria following?}
    Apart from criterion-level accuracy, we introduce three metrics to systematically assess a judge model’s
    (1) adherence to pluralistic criteria (PAcc),
    (2) sensitivity to criterion-level trade-offs within each prompt (TOS), and
    (3) ability to resolve criterion-level preference conflicts (CMR).
\end{enumerate}

Comprehensive evaluation across 25 LMMs exposes key challenges:
1) Strongest proprietary models (e.g., o4-mini, Claude-3.7-Sonnet) struggle to maintain consistent pluralistic criteria following, achieving 32.78\% in open-ended and 53.17\% in reasoning judgments;
2) open-source LMMs lag notably behind in both criterion-level accuracy and their ability to recognize trade-offs and preference conflicts; and
3) models fine-tuned on holistic, single-preference critic data primarily improve judgment consistency in visual grounding but generalize poorly to pluralistic, criterion-driven evaluation.
Additional analyses further reveals the boundaries of current LMM judges: 1) reasoning fine-tuning fails to enhance reasoning judgment but weakens models’ ability to recognize trade-offs; 2) test-time scaling effects are most evident in the strongest model, \textit{o4-mini}, but remain inconsistent across others in pluralistic criteria following; 3) the criterion-level upper bound of proprietary models aligns closely with human inter-annotator consistency, whereas open-source judges fail to exhibit this trend.

Overall, \ours\ is the first benchmark for evaluating multimodal judges under pluralistic criteria. Built on a carefully curated dataset and novel metrics, it reveals key limitations and informs more reliable, steerable evaluation.

\begin{figure*}[!ht]
    \centering
    \vspace{-1.9mm}
    \includegraphics[width=0.97\linewidth]{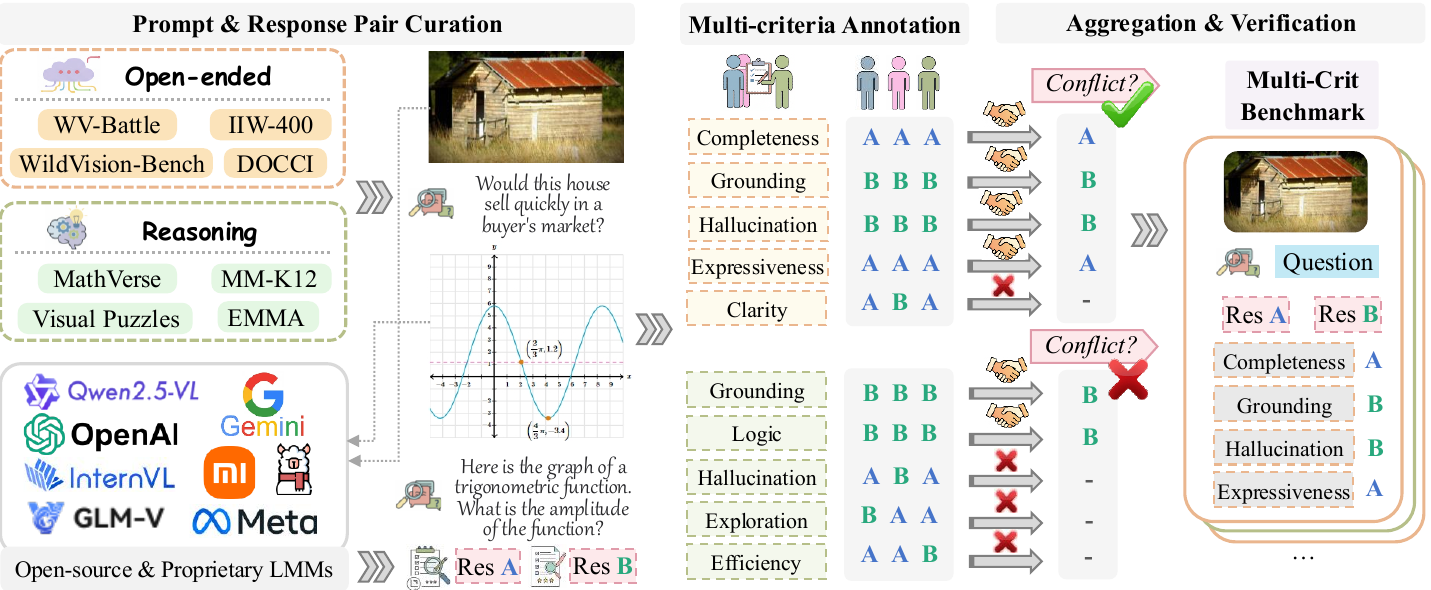}
    \vspace{-1.8mm}
    \caption{\textbf{Data Construction Pipeline.} \ours\ is built from diverse prompts across open-ended and reasoning tasks, responses from various LMMs reflecting subtle quality distinctions, and multi-criterion human annotations highlighting preference conflicts across criteria.}
    \label{fig:main-pipeline}
    \vspace{-4.7mm}
\end{figure*}

\vspace{-0.5mm}
\section{Related Work}
\vspace{-0.5mm}

\noindent\textbf{Large Multimodal Model as Judges and Critics.} Large multimodal models (LMMs) are widely adopted for evaluating model-generated responses, providing assessments ranging from scalar~\cite{li2024llava,yu2023mm,zhang2024mathverse,guan2024hallusionbench} to pairwise preference~\cite{liu2023visual,lu2024wildvision,yu2025rlaif}. 
GPT-4V~\cite{zhang2023gpt} was first shown to align closely with human judges across diverse multimodal evaluation tasks, motivating efforts to develop open-source counterparts. Subsequent works fine-tuned open-source LMMs into either BT-style reward models with dedicated reward heads~\cite{sun2023aligning, zang2025internlm, zhang2025basereward} or generative judges that produce textual justifications and scores~\cite{lee2024prometheus, xiong2025llava, zhang2025mm, zhang2025r1, wang2025unified, wang2025llava, unifiedreward-flex, unifiedreward-think, xiong2024llavaovchat,xiong2026phycritic} in an autoregressive manner. While our benchmark centers on generative judge models—due to their flexible instruction-following abilities inherent in the autoregressive generation process of LMMs—the proposed methodology is broadly applicable across diverse reward modeling paradigms.

\vspace{2mm}
\noindent\textbf{Benchmarking Multimodal Judges.}
Benchmarking reward and judge models is crucial for evaluating the reliability of AI feedback, which underpins both automated evaluation and the enhancement of LMM performance through preference alignment. 
MLLM-as-a-Judge~\cite{chen2024mllm} is the first to evaluate LMM as judges on academic tasks; VL-Rewardbench~\cite{li2025vl}, multimodal-rewardbench~\cite{yasunaga2025multimodal} and MM-RLHF~\cite{zhang2025mm} Benchs extend prompts sources to diverse real-world scenarios, reasoning and safety tasks. 
Following reward and judge benchmarks in text-only LLMs~\cite{lambert2024rewardbench, tan2024judgebench, zhou2024rmb, zheng2025learning}, these multimodal benchmarks evaluate judges by their overall preference between paired responses, measured against human or verifier-provided golden labels.
MM-Critic further extends evaluation to assess the correctness of generated textual critiques by leveraging GPT-generated scoring rubrics~\cite{zeng-etal-2025-mm}.
However, none of these benchmarks evaluate multimodal judges on their ability to follow diverse criteria for each response pair, and fine-grained, criterion-level evaluation remains underexplored.

\vspace{2mm}
\noindent\textbf{Criteria-Following Judgment.} When deploying model-based judges~\cite{zhang2024mathverse,yu2023mm, liu2023visual, yu2025rlaif, xiong2025llava, shirkavand2025cost,li2025groundingme} for automated evaluation, benchmarks usually define specific criteria such as helpfulness~\cite{lu2024wildvision}, visual grounding~\cite{ye2025painting}, or no hallucination~\cite{sun2023aligning,rawal2025argus}. While recent work began evaluating criterion-following of pure-text LLM judges—either by embedding criterion-level differences into responses~\cite{pitis2024improving} or summarizing from human justifications~\cite{wang2025rlbff}—we extend this line of research to the multimodal setting, analyzing how judges follow diverse criteria and capture preference conflicts.

\vspace{-0.3mm}
\section{\ours Benchmark}
\vspace{-0.5mm}

\ours is a  rigorously curated benchmark designed to evaluate the pluralistic criteria-following ability of multimodal judge models.
It includes diverse prompts from 8 data sources across open-ended and reasoning-dense domains, cross-model and intra-model response pairs from 11 LMMs of varying capabilities, and fine-grained multi-criteria human annotations within each prompt. The data construction pipeline is illustrated in Figure~\ref{fig:main-pipeline}. 
Additionally, three metrics are introduced in \ours to evaluate judge models on their ability to flexibly follow diverse evaluation criteria and identify criterion-level preference conflicts.

\subsection{Task Formulation}
\vspace{-0.5mm}

Existing multimodal judge benchmarks~\cite{li2025vl,yasunaga2025multimodal,zhang2025mm,hu2025multimodal} typically consist of preference pairs 
$(q, l_a, l_b, y)$, 
where $q$ is a multimodal prompt containing an image and a textual query, $l_a$ and $l_b$ are candidate responses, and $y \in \{a, b\}$ denotes the golden preference by human annotation or rule-based verification. 
However, such one-dimensional annotations cannot capture the multifaceted nature of human evaluation---different criteria may lead to conflicting judgments for the same response pair. 
\ours therefore extends this formulation to include criterion-specific preferences:
\[
(q, l_a, l_b, \{(c_i, y_i)\}_{i=1}^{K_q}),
\]
where each $c_i$ denotes the $i$-th evaluation criterion and $y_i \in \{a, b\}$ represents the preferred response under that criterion.

\renewcommand{\arraystretch}{0.92}  
\begin{figure*}[t]
\centering
\begin{minipage}[t]{0.37\textwidth}
    \vspace{-1.8mm}
    
    \small
    \setlength{\tabcolsep}{2pt}
    \centering
    \scalebox{0.9}{
    \begin{tabular}{l r}
    \toprule
    \textbf{Statistic} & \textbf{Number} \\
    \midrule
    \textit{Open-ended} &  \\
    - Prompts & 299 \\
    \quad - with conflicting preferences & 206 (68.9\%) \\
    - Criterion-level judgment & 1,000 \\
    \quad - Conflicting criterion pairs & 501 \\
    \midrule
    \textit{Verifiable Reasoning} & \\
    - Prompts & 126 \\
    \quad - with conflicting preferences &  109 (86.5\%) \\
    - Criterion-level judgment & 425 \\
    \quad - Conflicting criterion pairs & 281 \\
    \midrule
    Question length quantiles & (5, 10, 27) \\
    Response length quantiles & (104, 163.5, 247) \\
    \bottomrule
    \end{tabular}
    }
    \vspace{-2.2mm}
    \captionof{table}{\ours's key statistics.}
    \label{tab:mmcrit_stats_keys}
\end{minipage}
\hfill
\begin{minipage}[t]{0.60\textwidth}
    \centering
    \vspace{-2.3mm}
    \includegraphics[width=0.47\linewidth]{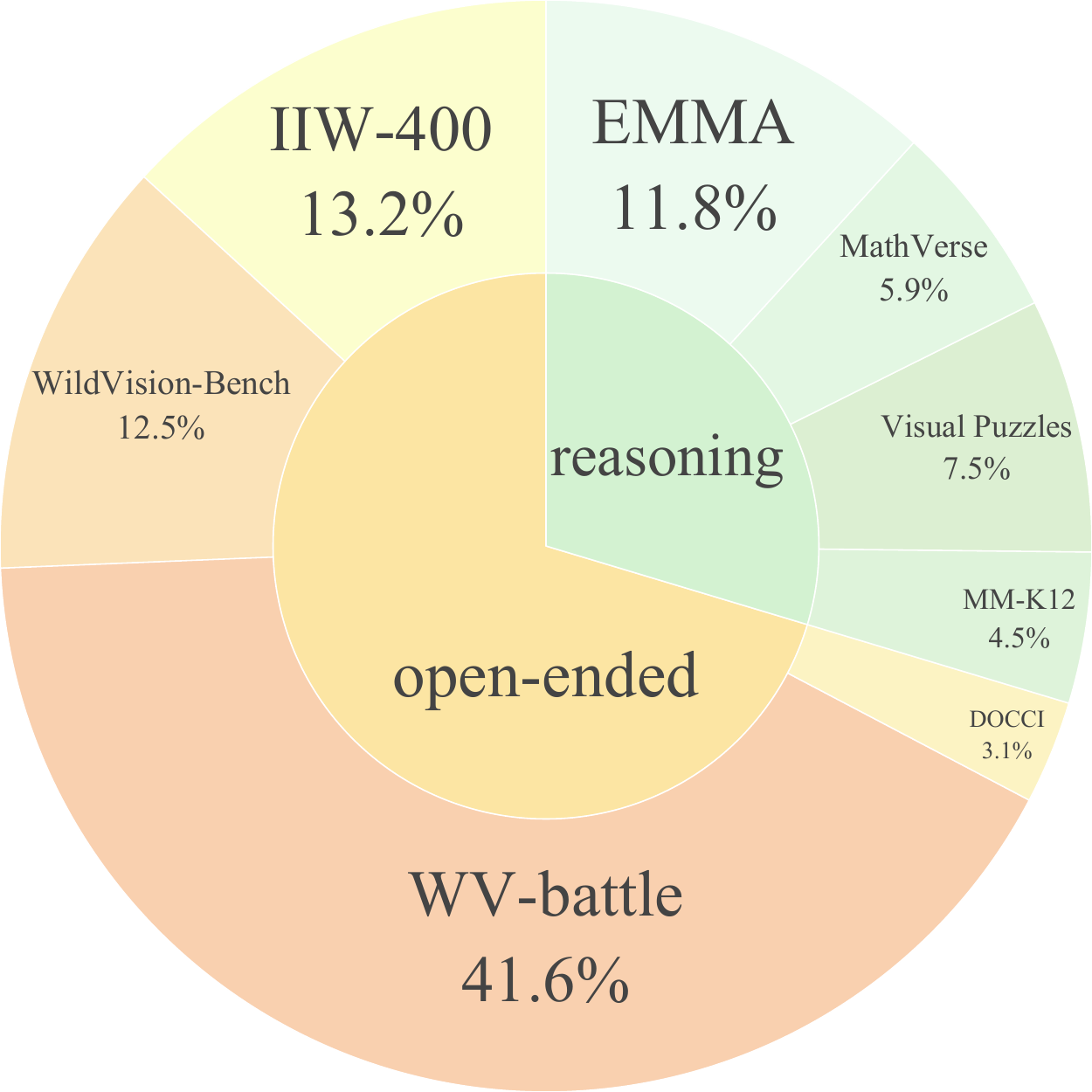}
    \hspace{1mm}
    \includegraphics[width=0.47\linewidth]{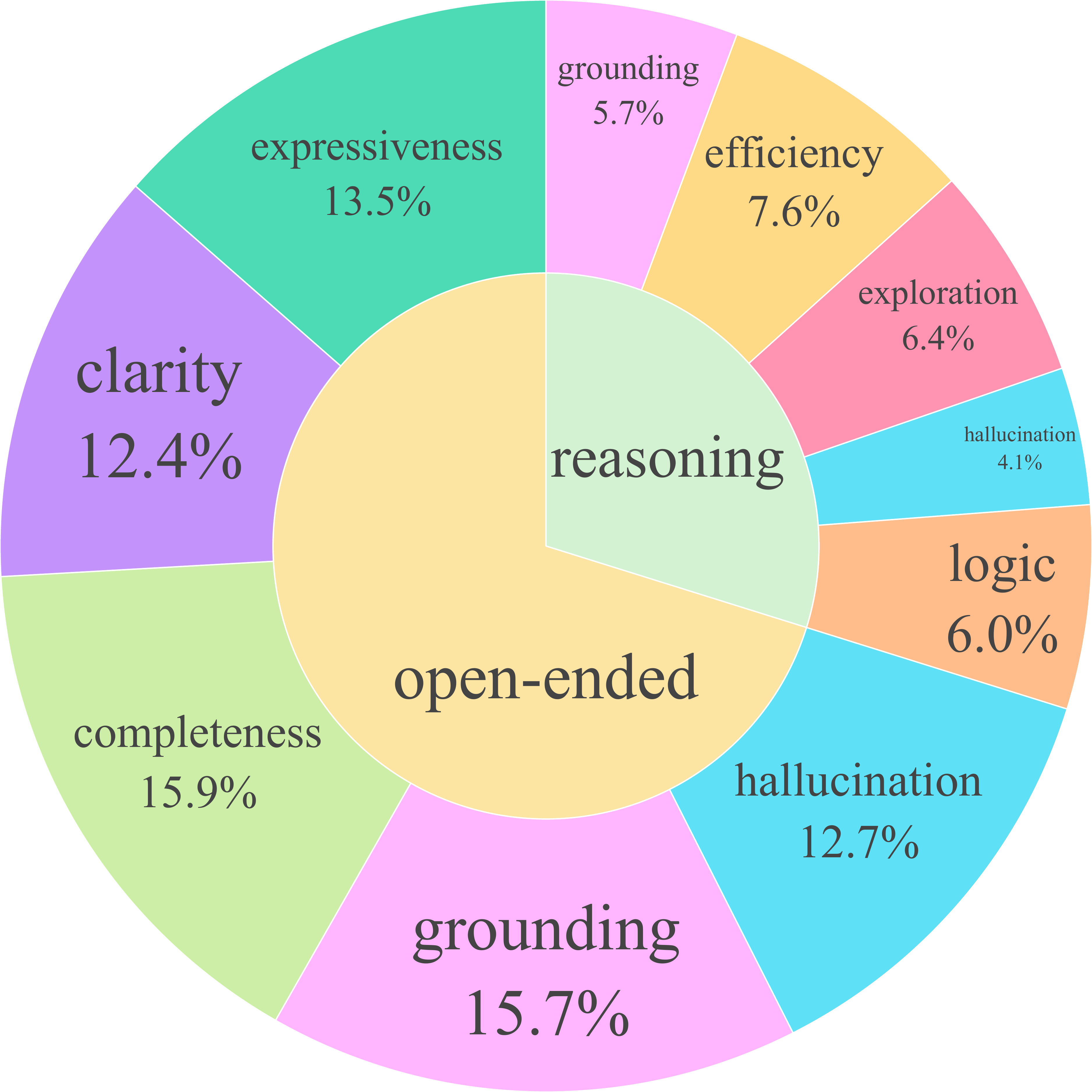}
    \vspace{-2.6mm}
    \caption{Distribution of prompt sources (left) and evaluation criteria (right).}
    \label{fig:data-distribution-circle}
\end{minipage}
\vspace{-5mm}
\end{figure*}
\renewcommand{\arraystretch}{1.0}  

\subsection{Prompt and Response Pair Curation}
\label{response-pair-curation}

Multi-Crit primarily covers two major scopes of multimodal judgment: 
(1) \textit{open-ended generation}, where responses are free-form and traditional fixed metrics are limited; and
(2) \textit{reasoning tasks}, where the judge model evaluates the quality of model-generated reasoning processes leading to objectively verifiable answers.

\vspace{2mm}
\noindent\textbf{Prompt Sources.} For open-ended generation, we select images from ImageInWords~\cite{garg2024imageinwords} and DOCCI~\cite{onoe2024docci} evaluation set for captioning and incorporate image–question pairs from the WildVision-Bench and -Battle datasets~\cite{lu2024wildvision}, which contain diverse, in-the-wild visual conversation queries.
For verifiable reasoning, we draw from reasoning-dense domains, including MathVerse~\cite{zhang2024mathverse} (math), MM-K12~\cite{meng2025mm} (STEM), EMMA-mini~\cite{hao2025can} (math, science, coding), and VisualPuzzles~\cite{song2025visualpuzzles} (visual analogy and puzzle solving).

 \vspace{2mm}
\noindent\textbf{Response Generation and Pair Curation.} To minimize data contamination, we generate candidate responses using 11 high-performing models, covering both proprietary (e.g. GPT-4o~\cite{hurst2024gpt}, Gemini-2.5-Flash~\cite{comanici2025gemini}) and open-source families (e.g., LLaMA3.2-V~\cite{grattafiori2024llama}, Qwen2.5-VL~\cite{bai2025qwen2}, InternVL3~\cite{chen2024internvl}); details in Appendix. We then form two types of response pairs: \textit{cross-model pairs}, generated by randomly selecting two different LMMs, and \textit{intra-model pairs}, produced by the same model via temperature sampling. For the latter, we sample $N=5$ responses at temperature 1.0 and select the pair with the largest cosine distance between their MiniLM-L6~\cite{wang2020minilm} embeddings to ensure content diversity. This process yields 3{,}538 response pairs.

\vspace{2mm}
\noindent\textbf{Filtering Challenging Pairs.} 
We apply a three-stage filtering mechanism to retain challenging yet balanced examples suitable for fine-grained evaluation. 
\textit{(1) Length normalization:} We exclude pairs with excessive length disparity, mostly retaining those within [0.7, 1.4] to mitigate superficial bias.
\textit{(2) Verifiable reasoning filtering:} For reasoning tasks, we use GPT-4o-mini to verify both responses and keep only pairs where both are correct or both incorrect, as correctness alone is trivial and our evaluation targets different criteria of reasoning quality.
\textit{(3) Ensemble-based difficulty filtering:} To emphasize nuanced criterion-related differences, we use three strong judge models (GPT-4o, Gemini-2.5-Flash, and Claude-3.7-Sonnet) for initial assessments of overall quality; pairs on which all judges unanimously agree are discarded, while the remaining ones are retained as challenging cases reflecting subtle criterion-level distinctions. A total of 707 response pairs are selected.

\subsection{Criteria-Specific Preference Annotation}

\noindent\textbf{Criteria Design.}
The criteria in \ours\ follow a clear set of principles to ensure balanced coverage of multimodal judging capabilities:
(1) \textit{Practicality}: reflecting common use cases of multimodal judges;
(2) \textit{Specificity}: capturing diverse quality aspects of generated responses while avoiding overlap; and
(3) \textit{Generality}: evaluating fundamental dimensions of model behavior rather than prompt-specific content.
Following these principles, we survey existing MLLM-as-a-Judge benchmarks, summarize their adopted criteria, and perform multi-round refinement to derive fivecriteria for open-ended tasks and another five for reasoning tasks, as briefed in Table~\ref{tab:criteria-summary} and detailed in Appendix.

\begin{table}[!h]
\centering
\vspace{-1.5mm}
\renewcommand{\arraystretch}{1.03}
\setlength{\tabcolsep}{2pt}
\scriptsize
\scalebox{1.03}{
\begin{tabular}{p{0.934\linewidth}}
\toprule
\texttt{\textbf{Criteria for Judging Open-ended Generation}} \\[-1pt]
\midrule 
\textbf{Completeness and Coverage:} Address the full scope of the task in the user's query, covering all major elements specified in the prompt as well as relevant visual aspects and contextual cues. \\[3pt]

\textbf{Visual Grounding and Details:} Reference observable elements in the image such as objects, spatial relationships, colors, or text, and bases its description or analysis on these details. \\[3pt]

\textbf{Factuality / No Hallucination:} Avoid visual or factual errors, ensuring all details and claims are presented in the image or reasonably supported by the prompt. \\[3pt]

\textbf{Creativity and Expressiveness:} Demonstrates imagination and originality when appropriate, or precise and knowledgeable articulation for analytical tasks, while remaining contextually appropriate. \\[3pt]

\textbf{Clarity and Coherence:} Communicates ideas clearly and logically, with fluent language, well-organized structure, and smooth flow of information. \\[1pt]

\midrule
\texttt{\textbf{Criteria for Judging Verifiable Reasoning}} \\[-1pt]
\midrule

\textbf{Visual Grounding:} Reference important visual elements—such as objects, layout, or text—and integrates them meaningfully into the reasoning. \\[3pt]

\textbf{Logic Coherence and Consistency:} Follow a clear, step-by-step logic without contradictions or unjustified leaps, and ensure the answer aligns with reasoning. \\[3pt]

\textbf{Factuality / No Hallucination:} Ensure accuracy of all claims and support them with the input, avoiding hallucinated visual details or factual errors. \\[3pt]

\textbf{Reflection and Exploration:} Demonstrate depth of reasoning through reflection and exploration of alternative interpretations, particularly in complex tasks. \\[3pt]

\textbf{Conciseness and Efficiency:} Remains concise and focused, matching the task complexity while avoiding redundancy or unnecessary over-analysis. \\[3pt]

\bottomrule
\end{tabular}
}
\vspace{-2.1mm}
\caption{Summary of the evaluation criteria employed in the \ours open-ended and reasoning splits, each capturing distinct dimensions of multimodal judgment.}
\label{tab:criteria-summary}
\vspace{-2.5mm}
\end{table}

\noindent\textbf{Rigorous Human Annotation.} We recruit nine computer science Ph.D. students as annotators, all with strong backgrounds in multimodal AI and STEM-related domains. Each receives detailed guidelines with clear illustrations of all evaluation criteria. During annotation, annotators are presented with a multimodal instance containing a prompt and two candidate responses. They exclusively evaluate one criterion at a time, determine which response is better (or declare a tie, limited to under 10\%), and provide a brief textual justification with supporting evidence. Annotators focus on response elements most relevant to the prompt and indicative of the given criterion.
To ensure consistency, we first establish a seed set of 10 open-ended and 10 reasoning examples. Annotators label this set individually, then participate in group discussions led by the project lead to align interpretations and reach consensus. After calibration, each evaluation sample—comprising the prompt, response pair, and set of criteria—is randomly assigned to three annotators for cross-validation in the main phase. The annotation process involved 289 hours of human effort in total. Inter-annotator agreement is quantified using the average pairwise Cohen’s $\kappa$~\cite{artstein2008inter}, achieving 0.718 on open-ended generation tasks and 0.805 on verifiable reasoning tasks, indicating substantial and reliable consistency among annotators.

\vspace{2mm}
\noindent\textbf{Preference Aggregation and Final Verification.} First, preferences for each criterion are aggregated, retaining only cases where all annotators agree on one model or where two annotators agree and the third declares a tie. Project leads manually review the textual justifications and discard samples with notable inconsistencies or verbose explanations. Subsequently, criterion-level judgments are aggregated at the prompt–response level. Finally, prompt-level filtering primarily retains pairs exhibiting criterion-level conflicts and balances the number of samples across all criteria.

\renewcommand{\arraystretch}{0.95}  
\setlength{\tabcolsep}{2pt}  
\definecolor{table-highlight}{HTML}{DDEEFF}  
\begin{table*}[!ht]
\adjustbox{max width=\textwidth}{
\begin{tabular}{l|ccc|ccccc|c}
\toprule
\multirow{2}{*}{Model}                 & \multicolumn{3}{c|}{Prompt-Level}                           & \multicolumn{6}{c}{Criterion-Level}                                          \\
\cline{2-4} \cline{5-10}
                                       & Pluralisic & Conflict & Tradeoff & Completeness &	Grounding	& Hallucination & Expressiveness & Clarity & Avg \\
\hline
\rowcolor{table-highlight}
\multicolumn{10}{c}{\textit{Open-Source LMMs}}\\
\hline

Qwen2.5-VL-7B-Instruct                 & 9.41  & 17.28 & 36.14 & 56.12 & 51.70  & 48.20  & 64.12 & 51.82 & 54.39 \\
LLaVA-OneVision-7B        & 11.46 & 12.77 & 37.86 & 53.20  & 48.80  & 43.10  & 55.68 & 52.26 & 50.61 \\

Llama-3.2-11B-Vision-Instruct          & 12.71 & 15.37 & 41.26 & 55.75 & 44.64 & 48.07 & 61.14 & 46.59 & 51.24 \\
Molmo-7B                  & 17.06 & 15.97 & 34.47 & 48.23 & 44.20  & 55.80  & 58.03 & 50.57 & 51.37 \\
Qwen3-VL-8B-Instruct             & 18.39 & 18.36 & 34.47 & 56.19 & 47.77 & 59.12 & 55.44 & 52.27 & 54.16 \\

MiniCPM-V-4.5-8B                   & 18.73 & 14.77 & 27.67 & 53.10  & 46.43 & 53.59 & 65.80  & 50.00    & 53.78 \\
Eagle2.5-8B                      & 20.40  & 23.75 & 50.97 & 64.16 & 54.91 & 52.49 & 59.59 & 54.55 & 57.14 \\
GLM-4.1V-9B-Thinking                   & 24.08 & 30.54 & 55.34 & 59.73 & 55.80  & 61.33 & 67.36 & 51.14 & 59.07 \\
InternVL3.5-8B                         & 25.08 & 32.34 & 62.14 & 61.50  & 59.38 & 56.35 & 69.43 & 58.52 & 61.04 \\
InternVL3-8B-Instruct                  & 26.09 & 30.34 & 61.17 & 64.60  & 62.95 & 56.35 & 67.88 & 56.25 & 61.61 \\
Qwen2.5-VL-72B-Instruct                & 28.43 & 35.53 & 60.68 & 69.47 & 63.39 & 58.56 & \second{70.98} & 56.82 & 63.84 \\

MiMo-VL-7B                    & 29.10  & 39.52 & 65.53 & 65.93 & 62.95 & 64.09 & 70.47 & 53.41 & 63.37 \\
InternVL3-78B-Instruct                 & 29.10  & 32.53 & 56.31 & \second{73.01} & 65.62 & 56.35 & 68.91 & 59.66 & 64.71 \\
InternVL3.5-38B                        & 30.43 & 33.73 & 64.08 & 71.68 & 65.18 & 62.98 & 63.73 & 61.93 & 65.10  \\

\hline
\rowcolor{table-highlight}
\multicolumn{10}{c}{\textit{Finetuned Judge Models}}\\

\hline
LLaVA-Critic-7B {\scriptsize(LlaVA-OV)}                     & 11.70 & 14.17 & 43.20 & 56.93 & 47.32 & 29.82 & 50.28 & 44.08 & 45.69 \\R1-Reward-7B {\scriptsize(Qwen2.5-VL)}                             & 17.73 & 20.36 & 45.63 & 59.29 & 60.71 & 49.72 & 55.44 & 53.98 & 55.83 \\
UnifiedReward-7B{\scriptsize(Qwen2.5-VL)}            & 18.06 & 8.38  & 16.50 & 57.96 & 52.23 & 52.49 & 57.51 & 55.68 & 55.17 \\
LLaVA-Critic-R1-7B{\scriptsize(Qwen2.5-VL)}                     & 18.39 & 17.96 & 39.32 & 55.31 & 57.59 & 46.96 & 63.73 & 55.11 & 55.74 \\

\hline
\rowcolor{table-highlight}
\multicolumn{10}{c}{\textit{Proprietary LMMs}}\\
\hline

Gemini-2.5-Flash                       & 25.42 & 31.34 & 62.14 & 64.60  & 60.71 & 66.30  & 63.73 & 57.39 & 62.55 \\
Gemini-2.5-Pro                         & 28.76 & 37.92 & \best{66.50}  & 65.93 & 59.82 & 70.17 & 62.18 & 60.23 & 63.67 \\
GPT-5                                  & 29.77 & 38.52 & 62.62 & 69.91 & 69.64 & \best{75.69} & 58.55 & \best{68.75} & 68.51 \\
o3                                     & 31.10 & 42.71 & 62.62 & 66.37 & \best{74.55} & \second{72.93} & 63.21 & \best{68.75} & 69.16 \\
GPT-4o                    & 31.44 & \best{44.91} & \second{66.02} & 68.14 & \second{71.88} & 65.75 & \best{76.17} & \second{65.91} & \second{69.57} \\
Claude-3.7-Sonnet             & \second{31.77} & 42.32 & 64.08 & 71.68 & \best{74.55} & 63.54 & 66.84 & 60.23 & 67.37 \\
o4-mini                                & \best{32.78} & \second{43.11} & 64.56 & \best{76.11} & \best{74.55} & 65.75 & 69.43 & 62.50  & \best{69.67} \\

\bottomrule
\end{tabular}
}
\vspace{-2mm}
\caption{\textbf{Results on open-ended split.} 
Criterion-level judgment performance is summarized by mean accuracy. The best and second best results are shown in \textbf{bold} and \underline{underlined} respectively. Notably, the top performing \textit{o4-mini} only achieves 32.78\% in pluralistic accuracy.}
\label{tab:table-open-ended}
\vspace{-4mm}
\end{table*}
\renewcommand{\arraystretch}{1.0}  

\subsection{Data Statistics}
\ours\ comprises 425 multimodal prompts and 1,425 criterion-level human judgments across open-ended and verifiable reasoning tasks, including 315 prompts with criterion-level conflicts that together constitute 782 conflicting criterion pairs. Key statistics and data distributions are shown in Table~\ref{tab:mmcrit_stats_keys} and Figure~\ref{fig:data-distribution-circle}, with details in the Appendix.

\subsection{Measuring Pluralistic Criteria Following}

\label{sec:methods-pluralistic-metrics}

We introduce three complementary metrics to evaluate how well multimodal judge models adhere to pluralistic criteria, identify criterion-level trade-offs, and capture preference conflicts within response pairs.

\vspace{1.8mm}
\noindent\textbf{Pluralistic Accuracy ($\text{PAcc}$).}  
Measures whether the judge produces correct judgments across \emph{all} criteria for each evaluation instance. 
Let $X$ denote the set of evaluation instances, where each $x = (q, l_a, l_b) \in X$ represents a multimodal prompt with its paired candidate responses, and let $C_x$ denote the set of evaluation criteria associated with $x$:
\[
\text{PAcc} = \frac{1}{|X|} \sum_{x \in X} \mathbb{I}\Bigg[ \bigwedge_{c \in C_x} \hat{y}_{x,c} = y_{x,c} \Bigg],
\]
where $\hat{y}_{x,c}$ denotes the judge’s predicted preference and $y_{x,c}$ the ground-truth label for criterion $c$.

\vspace{2mm}
\noindent\textbf{Trade-off Sensitivity (TOS).}  
Measures whether the judge model recognizes criterion-level trade-offs within each evaluation instance. Let $\mathcal{P}_x = \{\, (c_i, c_j) \in C_x \times C_x \mid y_{x,c_i} \neq y_{x,c_j} \,\}$  
denotes the set of criterion pairs in instance $x$ that exhibit ground-truth conflicts.  
\[
\text{TOS} = \frac{1}{|X^{*}|} \sum_{x \in X^{*}} \mathbb{I}\!\left[ \exists \,(c_i, c_j) \in \mathcal{P}_x : \hat{y}_{x,c_i} \neq \hat{y}_{x,c_j} \right],
\]
where $X^{*} = \{\, x \in X \mid \exists \ c_i, c_j \in C_x \text{ such that } y_{x,c_i} \neq y_{x,c_j} \,\}$  
denotes the subset of evaluation instances exhibiting at least one ground-truth conflict among criteria.  
TOS reflects whether the judge can \emph{perceive} criterion-level trade-offs within an instance, indicating its flexibility rather than absolute accuracy. A higher TOS suggests the model is less criterion-agnostic or overconfident in assigning identical preferences across all criteria.

\vspace{2mm}
\noindent\textbf{Conflict Matching Rate (CMR).}  
Measures the judge’s ability to correctly resolve ground-truth conflicts between criterion-based judgments. A conflict is considered \emph{matched} only if the model predicts both criteria in agreement with the ground truth preference:
{\small
\[
\text{CMR} =
\frac{
\sum_{x \in X^*} \sum_{(c_i,c_j)\in \mathcal{P}_x}
\mathbb{I}\!\left[(\hat{y}_{x,c_i}, \hat{y}_{x,c_j}) = (y_{x,c_i}, y_{x,c_j})\right]
}{
\sum_{x \in X^*} |\mathcal{P}_x|
}.
\]
}\noindent
This provides a \emph{stricter} and more fine-grained metric than TOS, evaluating whether the judge not only detects but also resolves conflicts consistently with humans.

\section{Experiments}

To comprehensively assess multimodal judge boundaries, we evaluate 25 LMMs across three model groups: (1) \textit{proprietary models}, including the GPT series~\cite{hurst2024gpt, jaech2024openai, openai2025o3o4mini}, Gemini family~\cite{comanici2025gemini}, and Claude-3.7-Sonnet~\cite{anthropic_claude3.7_sonnet_2025}; (2) \textit{leading open-source LMMs}, including Qwen2.5/3-VL~\cite{bai2025qwen2, yang2025qwen3}, InternVL3/3.5~\cite{zhu2025internvl3, wang2025internvl3.5}, GLM-4.1V-Thinking~\cite{hong2025glm}, LLaMA3.2-V~\cite{grattafiori2024llama}, MiniCPM-V-4.5~\cite{yu2025minicpm}, Eagle2.5~\cite{chen2025eagle}, Molmo~\cite{deitke2025molmo} and LLaVA-OneVision~\cite{li2024llava}; and (3) \textit{finetuned judge models}, including LLaVA-Critic~\cite{xiong2025llava}/-R1~\cite{wang2025llava}, R1-Reward~\cite{zhang2025r1} and UnifiedReward~\cite{wang2025unified}.

\vspace{2mm}
\noindent\textbf{Implementation Details.} During inference, each model was prompted to evaluate a single multimodal sample containing one prompt and a pair of responses, focusing exclusively on one criterion at a time. The full evaluation prompts are provided in the Appendix, with minor adjustments for model-specific formatting. Sampling temperatures follow the original repository settings, or default to 0.6 with a top-$p$ of 0.95 when unspecified.

\renewcommand{\arraystretch}{0.94}
\setlength{\tabcolsep}{4pt}  
\begin{table*}[!ht]
\vspace{-1.5mm}
\adjustbox{max width=\textwidth}{
\begin{tabular}{l|ccc|ccccc|c}
\toprule
\multirow{2}{*}{Model}                 & \multicolumn{3}{c|}{Prompt-Level}                           & \multicolumn{6}{c}{Criterion-Level}                                          \\
\cline{2-4} \cline{5-10}
                                       & Pluralisic & Conflict & Tradeoff & Grounding & Logic & Hallucination & Exploration & Efficiency & Avg \\
\rowcolor{table-highlight}
\hline
\multicolumn{10}{c}{\textit{Open-Source LMMs}}\\
\hline

LLaVA-OneVision-7B         & 9.52            & 17.08             & 51.38                & 40.74        & 44.19 & 29.31         & 50.55       & 58.72       & 44.70     \\
Llama-3.2-11B-Vision-Instruct          & 10.32           & 19.93             & 44.95                & 49.38        & 45.35 & 43.10          & 53.85       & 62.39       & 50.81     \\
Molmo-7B                  & 11.11           & 13.17             & 34.86                & 45.68        & 40.70 & 43.10          & 43.96       & 61.47       & 46.98     \\
Qwen3-VL-8B-Instruct             & 16.67           & 19.22             & 35.78                & 50.62        & 56.98 & 65.52         & 47.25       & 64.22       & 56.92     \\
Qwen2.5-VL-7B-Instruct                 & 16.67           & 25.62             & 66.06                & 46.91        & 52.33 & 55.17         & 57.14       & 65.14       & 55.34     \\
Eagle2.5-8B                      & 19.05           & 27.05             & 48.62                & 49.38        & 47.67 & 44.83         & 58.24       & 69.72       & 53.97     \\
GLM-4.1V-9B-Thinking                   & 23.02           & 27.40              & 55.05                & 54.32        & 60.47 & 65.52         & 50.55       & 66.97       & 59.57     \\
MiniCPM-V-4.5-8B           & 25.40            & 38.08             & 75.23                & 56.79        & 66.28 & 56.90          & 65.93       & 66.97       & 62.57     \\
InternVL3-8B-Instruct                  & 26.98           & 39.50              & 66.06                & 64.20         & 65.12 & 58.62         & 67.03       & 76.15       & 66.22     \\
InternVL3-78B-Instruct                 & 29.37           & 39.50              & 67.89                & 62.96        & 69.77 & 53.45         & 63.74       & 74.31       & 64.85     \\
Qwen2.5-VL-72B-Instruct                & 32.54           & 45.91             & 77.06                & 60.49        & 61.63 & 56.90          & 68.13       & 75.23       & 64.48     \\
InternVL3.5-8B                         & 32.54           & 39.15             & 69.72                & 65.43        & 63.95 & 56.90          & 71.43       & 71.56       & 65.85     \\
MiMo-VL-7B    & 37.30            & 41.99             & 71.56                & 59.26        & 59.30  & 70.69         & 67.03       & 75.23       & 66.30      \\
InternVL3.5-38B                        & 37.30           & 47.69             & 75.23                & 71.60        & \underline{70.93} & 55.17         & 78.02       & 73.39       & 69.82     \\

\hline
\rowcolor{table-highlight}
\multicolumn{10}{c}{\textit{Finetuned Judges}}\\
\hline

UnifiedReward-7B {\scriptsize(Qwen2.5-VL)}            & 11.90            & 13.17             & 22.02                & 64.20         & \underline{70.93} & 50.00            & 58.24       & 54.13       & 59.50
\\LLaVA-Critic-7B {\scriptsize(LLaVA-OV)}                    & 13.49           & 22.42             & 50.46                & 43.21        & 44.19 & 43.10          & 45.05       & 60.55       & 47.22     \\

LLaVA-Critic-R1-7B{\scriptsize(Qwen2.5-VL)}                      & 16.67           & 16.37             & 49.54                & 53.09        & 50.00    & 51.72         & 56.04       & 55.05       & 53.18     \\
R1-Reward    {\scriptsize(Qwen2.5-VL)}                           & 19.05           & 24.56             & 62.39                & 58.02        & 54.65 & 46.55         & 52.75       & 60.55       & 54.50      \\

\hline
\rowcolor{table-highlight}
\multicolumn{10}{c}{\textit{Proprietary Judge Models}}\\
\hline

Claude-3.7-Sonnet           & 36.51 & 51.96 & \second{83.49} & 74.07 & 62.79 & 65.52 & 68.13 & 76.15 & 69.33 \\
Gemini-2.5-Flash                       & 37.30 & 53.05 & 73.39 & 67.50 & 63.95 & 64.91 & 78.02 & 77.98 & 70.47 \\
Gemini-2.5-Pro                        & 41.27 & 52.33 & 75.93 & \best{79.01} & 67.44 & 68.97 & 75.56 & 74.31 & 73.06 \\
GPT-4o                     & 41.27 & 55.16 & \best{84.40} & 61.73 & 67.44 & 55.17 & 80.22 & \second{84.40} & 69.79 \\
o3                                     & 44.44 & \second{62.28} & 82.57 & 67.90 & \second{70.93} & \best{84.21} & \best{84.62} & 81.65 & \second{77.86} \\
GPT-5                                  & \second{45.24} & 56.58 & 78.90 & 74.07 & \second{70.93} & \second{82.76} & 81.32 & 77.98 & 77.41 \\
o4-mini & \best{53.17} & \best{65.84} & \second{83.49} & \second{77.78} & \best{75.58} & 79.31 & \second{83.52} & \best{88.07} & \best{80.85} \\
\bottomrule
\end{tabular}
}
\vspace{-2mm}
\caption{\textbf{Results on verifiable reasoning.} Best and second-best results are in \textbf{bold} and \underline{underlined}. }
\vspace{-4mm}
\label{tab:table-reasoning}
\end{table*}
\renewcommand{\arraystretch}{1.0}

\subsection{Main Results}
Results are presented in Table~\ref{tab:table-open-ended} for open-ended responses and Table~\ref{tab:table-reasoning} for reasoning responses, revealing the following key observations and analyses:

\begin{figure}[!tp]
    \centering
    \includegraphics[width=0.99\linewidth]{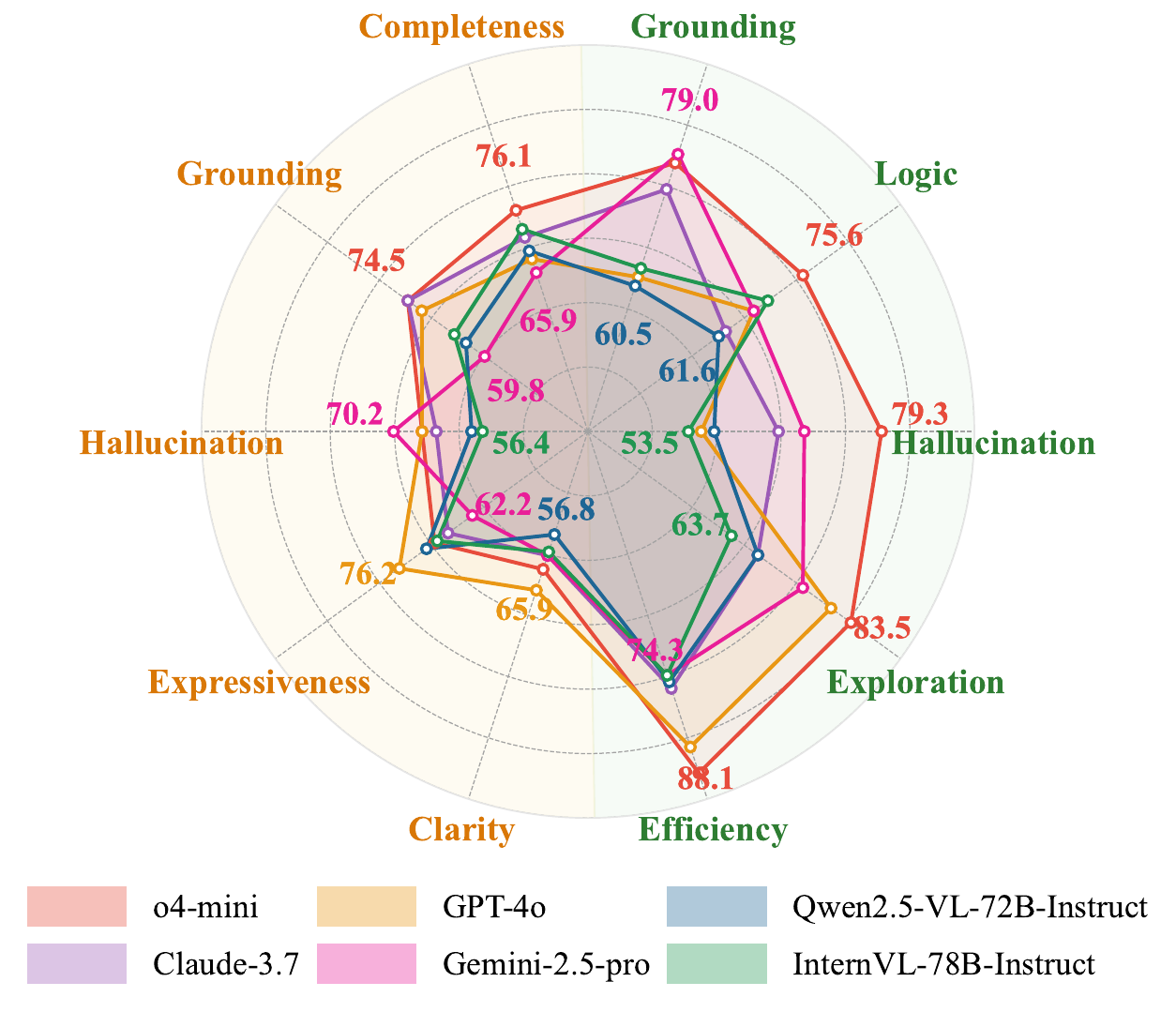}
    \vspace{-1.5mm}
    \caption{
    Average performance across each criterion. While the top model differs across criteria, all models show stronger pluralistic alignment in \textcolor[HTML]{2E7D32}{verifiable reasoning} than in \textcolor[HTML]{D97706}{open-ended} tasks.
    }
    \vspace{-6.5mm}
    \label{fig:lidar-results}
\end{figure}

\vspace{2mm}
\noindent\textbf{Pluralistic judging is highly challenging, especially for open-ended tasks.} The Pluralistic Accuracy (PAcc), which requires correct preferences across all criteria for a single query, is notably low. 
\textit{o4-mini} and \textit{Claude-3.7-Sonnet} achieve the highest open-ended PAcc at 32.78\% and 31.77\%, while \textit{o4-mini} and \textit{GPT-5} lead on verifiable reasoning with 53.17\% and 45.24\%. This highlights the difficulty of pluralistic judging even for state-of-the-art LMMs.
As in Figure~\ref{fig:lidar-results}, for top-performing models, results on open-ended tasks are lower, reflecting the greater subjectivity and reliance on fine-grained visual perception for evaluating free-form responses (e.g., real-world images or screenshots). In contrast, existing multimodal reasoning tasks—typically involving multimodal math, STEM, puzzle, and coding problems—are more objective, with clearer verification and explicit visual grounding.
These findings highlight the inherent difficulty of pluralistic judgment and underscore the necessity of the proposed \ours\ benchmark.

\vspace{2mm}
\noindent\textbf{Model performance varies across criteria and metrics.}
For judgment in verifiable reasoning, \textit{o4-mini} performs best on Logic (75.58\%) and Efficiency (88.07\%), but is surpassed by \textit{o3} in Hallucination (84.21\%) and Exploration (84.62\%), and by \textit{Gemini-2.5-Pro} in Grounding (79.01\%).
At the prompt level, \textit{GPT-4o} shows the highest sensitivity to trade-offs in prompts with criterion conflicts, achieving the best Trade-off Sensitivity of 84.40\%. However, its ability to resolve conflicting criterion-level preference pairs is notably weaker than that of the top-performing \textit{o4-mini} (55.16\% vs. 65.84\% in CMR). 
A similar trend is observed in open-ended tasks: \textit{GPT-4o-} excels in \textit{Expressiveness} (76.17\%) but lags in \textit{Completeness} (68.14\%) and \textit{Hallucination} (65.75\%). In contrast, \textit{o3} performs strongly on \textit{Hallucination} (72.93\%) but ranks low in \textit{Expressiveness} (63.21\%).
These results highlight the multifaceted nature of pluralistic judgment—no model excels across all aspects, each showing distinct strengths and blind spots.

\vspace{2mm}
\noindent\textbf{Open-source LMMs fall short in pluralistic criterion-following, especially in capturing criterion-level preference conflicts.} As shown in Tables~\ref{tab:table-open-ended} and~\ref{tab:table-reasoning}, proprietary LMMs (e.g., \textit{o4-mini}, \textit{GPT-5}, \textit{o3}) maintain a clear lead over open-source counterparts (e.g., \textit{InternVL3.5-38B}, \textit{Qwen2.5-VL-72B-Instruct}) across all prompt-level metrics (PAcc, CMR, and TOS) and criterion-level accuracies.
Specifically, the conflict matching rate drops by about 9.4 points in open-ended tasks (43.1~$\rightarrow$~33.7) and 18.1 points in reasoning tasks (65.8~$\rightarrow$~47.7), far exceeding the 4- and 11-point declines observed in criterion-level accuracy.
This indicates that open-source models struggle to maintain consistent pluralistic evaluation across conflicting criteria, highlighting their limited ability to integrate multi-faceted criterion-level signals into reliable judgments.

\vspace{2mm}
\noindent\textbf{Finetuned judge models improve evaluation in visual grounding but are limited in handling pluralistic criteria.} A striking observation is that models specifically fine-tuned as judges or reward models (e.g., \textit{LLaVA-Critic}~\cite{xiong2025llava}, \textit{R1-Reward}~\cite{zhang2025r1}, \textit{LLaVA-Critic-R1}~\cite{wang2025llava}, \textit{UnifiedReward-Think}~\cite{wang2025unified}) do not exhibit consistent performance gains over their base models, unlike prior benchmarks~\cite{li2025vl,tan2024judgebench}.
For Qwen-based fine-tuned judges, \textit{R1-Reward} is the only model that outperforms the base \textit{Qwen2.5-VL-7B-Instruct} in \textit{PAcc} across both  the open-ended split (9.41$\rightarrow$17.73) and  the reasoning split (16.67$\rightarrow$19.05)—yet its improvement 
at recognizing trade-off (36.14$\rightarrow$45.63, 66.06$\rightarrow$62.39)
and conflict capturing (17.28$\rightarrow$20.36, 25.62$\rightarrow$24.56) remains inconsistent. Other fine-tuned judges perform similarly or worse than similar-sized models without critic tuning, remaining far behind proprietary counterparts.

Another notable trend is that all Qwen-based fine-tuned judges show consistent improvements in evaluating visual grounding across both splits—
\textit{R1-Reward} (51.7$\rightarrow$60.7, 46.9$\rightarrow$58.0),
\textit{LLaVA-Critic-R1} (51.7$\rightarrow$57.6, 46.9$\rightarrow$53.1), and
\textit{UnifiedReward} (51.7$\rightarrow$52.2, 46.9$\rightarrow$64.2).
For the LLaVA-based judge, \textit{LLaVA-Critic-7B (LLaVA-OV)} improves on reasoning (40.7$\rightarrow$43.2) but a marginal drop on open-ended evaluation (48.8$\rightarrow$47.3). 

Overall, current critic fine-tuning pipelines—driven by holistic preference signals and limited critic prompt templates—mainly enhance judgment capacity in visual–textual grounding and alignment, but fail to generalize to diverse, fine-grained, and often conflicting evaluation criteria, thereby underscoring the need for pluralistic, criterion-aware training as introduced in \ours.

\vspace{-0.5mm}
\subsection{Additional Discussions}
\label{subsec:experiment-additional-discussion}
\begin{figure}[!h]
    \centering
    \vspace{-2.8mm}
    \includegraphics[width=1\linewidth]{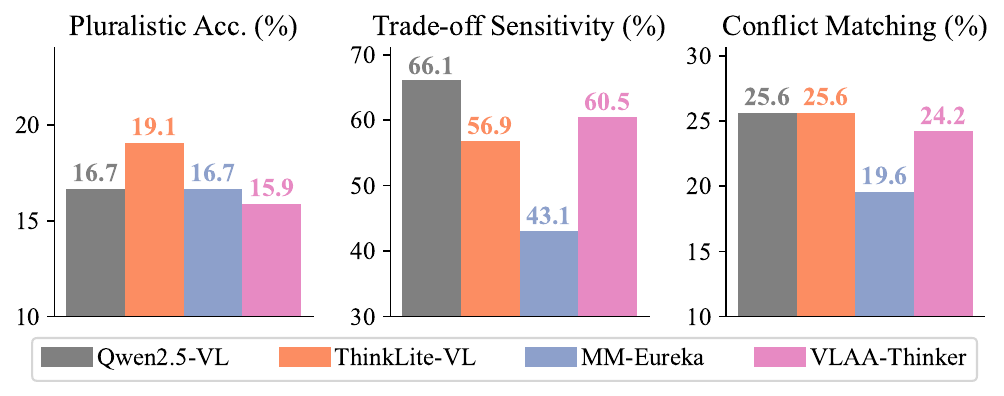}
    \vspace{-7.7mm}
    \caption{Results of RL-tuned reasoning models on the \ours\ reasoning split, all based on Qwen2.5-VL-7B.}
    \vspace{-4.6mm}
    \label{fig:results-reasoning-finetuned-reasoning-models}
\end{figure}

\vspace{-0.5mm}
To further investigate the capabilities and limitations of LMM judges in pluralistic criteria-following, we conduct three additional analyses as follows:

\vspace{1.8mm}
\noindent\textbf{1. Does RLVR on multimodal reasoning improve judgment in reasoning traces?} We evaluate three GRPO-finetuned models~\cite{wang2025sota, meng2025mm, chen2025sftrlearlyinvestigation} on general reasoning tasks (e.g., math and charts), all showing clear reasoning gains. However, these models, trained on holistic accuracy rewards, show reduced sensitivity to trade-offs and conflict matching in reasoning judgments (Figure~\ref{fig:results-reasoning-finetuned-reasoning-models}), indicating weaker recognition of criterion-level preference conflicts.

\begin{figure}[!h]
    \centering
    \vspace{-3.2mm}
    \includegraphics[width=0.98\linewidth]{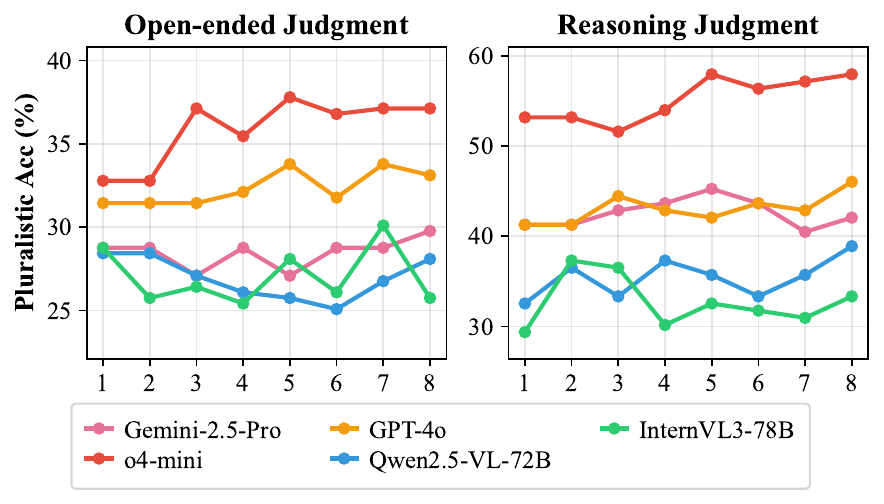}
    \vspace{-3.5mm}
    \caption{Test-time scaling behavior by pluralistic accuracy.}
    \vspace{-4mm}
    \label{fig:test-time-scaling}
\end{figure}

\vspace{2mm}
\noindent\textbf{2. Does test-time scaling hold?}
We run each judge model $K$ times, take a majority vote for each criterion sample as the final judgment, and then compute the pluralistic accuracy.
As shown in Figure~\ref{fig:test-time-scaling}, the strongest model, \textit{o4-mini}, exhibits the most robust test-time scaling behavior for both open-ended (32.78$\rightarrow$37.12) and reasoning (53.17$\rightarrow$57.94) judgments, with \textit{GPT-4o} showing similar gains. Other models show inconsistent trends and large variance, indicating that test-time scaling offers only limited and model-dependent benefits in multi-criteria judgment.

\begin{figure}[!h]
    \centering
    \vspace{-2mm}
    \includegraphics[width=1\linewidth]{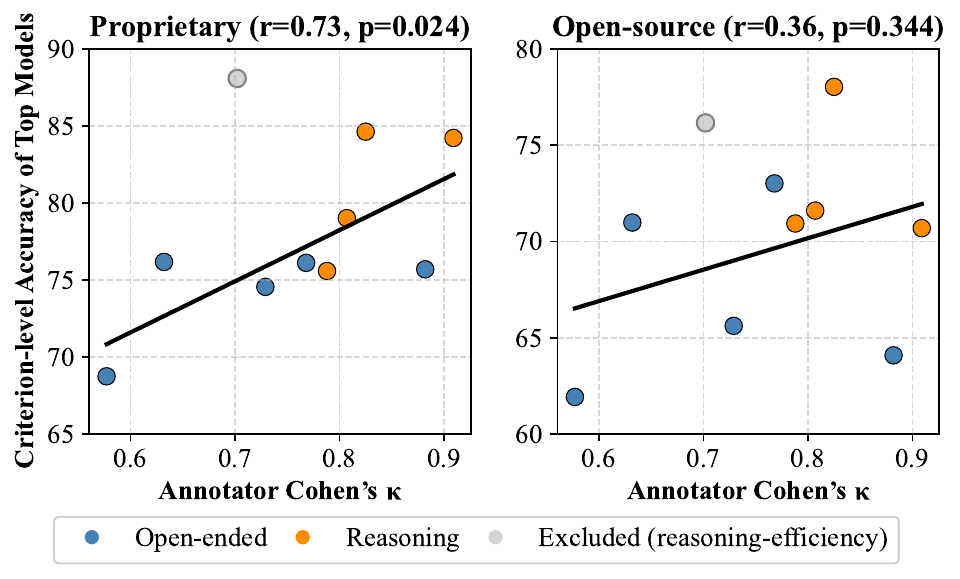}
    \vspace{-6.5mm}
    \caption{Correlation between criterion-level top judge accuracy and inter-annotator agreement.
Each point denotes one criterion.}
    \label{fig:criterion-level-acc-interhuman-annotator}
    \vspace{-4.5mm}
\end{figure}

\vspace{2mm}
\noindent\textbf{3. What are the current boundaries of criteria following for open-source and proprietary models?} To probe the upper limits of criterion-level consistency, we correlate each model group’s strongest criterion accuracies with human inter-annotator agreement (Cohen’s~$\kappa$), excluding \textit{reasoning-efficiency}, which is largely influenced by response length. As in Figure~\ref{fig:criterion-level-acc-interhuman-annotator}, proprietary models exhibit a strong and significant correlation (r=0.73, p=0.024), indicating that their upper-bound judgment patterns align closely with human agreement. In contrast, open-source models show a weaker, more inconsistent relationship (r=0.36, p=0.344), reflecting limited capacity to internalize human-consistent evaluation criteria. These findings suggest that while open-source models would benefit from scaling high-quality human annotations to improve pluralistic criteria-following, proprietary systems might face the next challenge of surpassing human-level evaluation alignment.

\section{Conclusion}

We introduced \ours, a comprehensive benchmark for evaluating multimodal judges under pluralistic, criterion-level settings. By providing fine-grained multi-criterion human annotations and complementary pluralistic metrics, \ours\ enables systematic analysis of models’ adherence to diverse criteria, trade-off recognition, and conflict resolution. Comprehensive experiments reveal key limitations of existing \textit{LMM-as-a-Judge} systems: proprietary models struggle with consistent pluralistic adherence, particularly in open-ended evaluation; open-source models lag further behind in their criteria-following capacities; and critic fine-tuning, while improving visual grounding and reference alignment, remains narrowly effective and fails to generalize to fine-grained or conflicting evaluation dimensions. Additional analyses highlight diminished trade-off recognition in reasoning fine-tuning, inconsistent test-time scaling behaviors, and the human-aligned upper bounds of proprietary models in criterion-level consistency.
In summary, \ours\ serves as a challenging suite to explore the boundaries of current AI-based multimodal judge systems, paving the way toward more steerable, reliable, and even superhuman AI feedback.

\paragraph{Acknowledgement.} We thank Chunyuan Li for his valuable feedback on the project idea. This work was partially supported by NSF IIS 2347592, 2348169, DBI 2405416, CCF 2348306, CNS 2347617, RISE 2536663.
{
    \small
    \bibliographystyle{ieeenat_fullname}
    \bibliography{main}

\begin{thebibliography}{73}
\providecommand{\natexlab}[1]{#1}
\providecommand{\url}[1]{\texttt{#1}}
\expandafter\ifx\csname urlstyle\endcsname\relax
  \providecommand{\doi}[1]{doi: #1}\else
  \providecommand{\doi}{doi: \begingroup \urlstyle{rm}\Url}\fi

\bibitem[Artstein and Poesio(2008)]{artstein2008inter}
Ron Artstein and Massimo Poesio.
\newblock Inter-coder agreement for computational linguistics.
\newblock \emph{Computational linguistics}, 34\penalty0 (4):\penalty0 555--596, 2008.

\bibitem[Bai et~al.(2025)Bai, Chen, Liu, Wang, Ge, Song, Dang, Wang, Wang, Tang, et~al.]{bai2025qwen2}
Shuai Bai, Keqin Chen, Xuejing Liu, Jialin Wang, Wenbin Ge, Sibo Song, Kai Dang, Peng Wang, Shijie Wang, Jun Tang, et~al.
\newblock Qwen2. 5-vl technical report.
\newblock \emph{arXiv preprint arXiv:2502.13923}, 2025.

\bibitem[Chen et~al.(2024{\natexlab{a}})Chen, Chen, Zhang, Wang, Liu, Zhou, Zhang, Wan, Zhou, and Sun]{chen2024mllm}
Dongping Chen, Ruoxi Chen, Shilin Zhang, Yaochen Wang, Yinuo Liu, Huichi Zhou, Qihui Zhang, Yao Wan, Pan Zhou, and Lichao Sun.
\newblock Mllm-as-a-judge: Assessing multimodal llm-as-a-judge with vision-language benchmark.
\newblock In \emph{Forty-first International Conference on Machine Learning}, 2024{\natexlab{a}}.

\bibitem[Chen et~al.(2025{\natexlab{a}})Chen, Li, Wang, Jiang, Liu, Lu, Huang, Byeon, Le, Rintamaki, et~al.]{chen2025eagle}
Guo Chen, Zhiqi Li, Shihao Wang, Jindong Jiang, Yicheng Liu, Lidong Lu, De-An Huang, Wonmin Byeon, Matthieu Le, Tuomas Rintamaki, et~al.
\newblock Eagle 2.5: Boosting long-context post-training for frontier vision-language models.
\newblock \emph{arXiv preprint arXiv:2504.15271}, 2025{\natexlab{a}}.

\bibitem[Chen et~al.(2025{\natexlab{b}})Chen, Tu, Wang, Liu, Tang, Du, Zhou, and Xie]{chen2025sftrlearlyinvestigation}
Hardy Chen, Haoqin Tu, Fali Wang, Hui Liu, Xianfeng Tang, Xinya Du, Yuyin Zhou, and Cihang Xie.
\newblock Sft or rl? an early investigation into training r1-like reasoning large vision-language models, 2025{\natexlab{b}}.

\bibitem[Chen et~al.(2023)Chen, Xiong, Wu, Liu, Hu, Chen, Chen, Liu, and Huang]{chen2023gpt}
Ruibo Chen, Tianyi Xiong, Yihan Wu, Guodong Liu, Zhengmian Hu, Lichang Chen, Yanshuo Chen, Chenxi Liu, and Heng Huang.
\newblock Gpt-4 vision on medical image classification--a case study on covid-19 dataset.
\newblock \emph{arXiv preprint arXiv:2310.18498}, 2023.

\bibitem[Chen et~al.(2024{\natexlab{b}})Chen, Wu, Wang, Su, Chen, Xing, Zhong, Zhang, Zhu, Lu, et~al.]{chen2024internvl}
Zhe Chen, Jiannan Wu, Wenhai Wang, Weijie Su, Guo Chen, Sen Xing, Muyan Zhong, Qinglong Zhang, Xizhou Zhu, Lewei Lu, et~al.
\newblock Internvl: Scaling up vision foundation models and aligning for generic visual-linguistic tasks.
\newblock In \emph{Proceedings of the IEEE/CVF conference on computer vision and pattern recognition}, pages 24185--24198, 2024{\natexlab{b}}.

\bibitem[Comanici et~al.(2025)Comanici, Bieber, Schaekermann, Pasupat, Sachdeva, Dhillon, Blistein, Ram, Zhang, Rosen, et~al.]{comanici2025gemini}
Gheorghe Comanici, Eric Bieber, Mike Schaekermann, Ice Pasupat, Noveen Sachdeva, Inderjit Dhillon, Marcel Blistein, Ori Ram, Dan Zhang, Evan Rosen, et~al.
\newblock Gemini 2.5: Pushing the frontier with advanced reasoning, multimodality, long context, and next generation agentic capabilities.
\newblock \emph{arXiv preprint arXiv:2507.06261}, 2025.

\bibitem[Deitke et~al.(2025)Deitke, Clark, Lee, Tripathi, Yang, Park, Salehi, Muennighoff, Lo, Soldaini, et~al.]{deitke2025molmo}
Matt Deitke, Christopher Clark, Sangho Lee, Rohun Tripathi, Yue Yang, Jae~Sung Park, Mohammadreza Salehi, Niklas Muennighoff, Kyle Lo, Luca Soldaini, et~al.
\newblock Molmo and pixmo: Open weights and open data for state-of-the-art vision-language models.
\newblock In \emph{Proceedings of the Computer Vision and Pattern Recognition Conference}, pages 91--104, 2025.

\bibitem[Garg et~al.(2024)Garg, Burns, Karagol-Ayan, Bitton, Montgomery, Onoe, Bunner, Krishna, Baldridge, and Soricut]{garg2024imageinwords}
Roopal Garg, Andrea Burns, Burcu Karagol-Ayan, Yonatan Bitton, Ceslee Montgomery, Yasumasa Onoe, Andrew Bunner, Ranjay Krishna, Jason~Michael Baldridge, and Radu Soricut.
\newblock Imageinwords: Unlocking hyper-detailed image descriptions.
\newblock In \emph{Proceedings of the 2024 Conference on Empirical Methods in Natural Language Processing}, pages 93--127, 2024.

\bibitem[Grattafiori et~al.(2024)Grattafiori, Dubey, Jauhri, Pandey, Kadian, Al-Dahle, Letman, Mathur, Schelten, Vaughan, et~al.]{grattafiori2024llama}
Aaron Grattafiori, Abhimanyu Dubey, Abhinav Jauhri, Abhinav Pandey, Abhishek Kadian, Ahmad Al-Dahle, Aiesha Letman, Akhil Mathur, Alan Schelten, Alex Vaughan, et~al.
\newblock The llama 3 herd of models.
\newblock \emph{arXiv preprint arXiv:2407.21783}, 2024.

\bibitem[Guan et~al.(2024)Guan, Liu, Wu, Xian, Li, Liu, Wang, Chen, Huang, Yacoob, et~al.]{guan2024hallusionbench}
Tianrui Guan, Fuxiao Liu, Xiyang Wu, Ruiqi Xian, Zongxia Li, Xiaoyu Liu, Xijun Wang, Lichang Chen, Furong Huang, Yaser Yacoob, et~al.
\newblock Hallusionbench: an advanced diagnostic suite for entangled language hallucination and visual illusion in large vision-language models.
\newblock In \emph{Proceedings of the IEEE/CVF Conference on Computer Vision and Pattern Recognition}, pages 14375--14385, 2024.

\bibitem[Hao et~al.(2025)Hao, Gu, Wang, Li, Yang, Wang, and Cheng]{hao2025can}
Yunzhuo Hao, Jiawei Gu, Huichen~Will Wang, Linjie Li, Zhengyuan Yang, Lijuan Wang, and Yu Cheng.
\newblock Can mllms reason in multimodality? emma: An enhanced multimodal reasoning benchmark.
\newblock \emph{arXiv preprint arXiv:2501.05444}, 2025.

\bibitem[Hong et~al.(2025)Hong, Yu, Gu, Wang, Gan, Tang, Cheng, Qi, Ji, Pan, et~al.]{hong2025glm}
Wenyi Hong, Wenmeng Yu, Xiaotao Gu, Guo Wang, Guobing Gan, Haomiao Tang, Jiale Cheng, Ji Qi, Junhui Ji, Lihang Pan, et~al.
\newblock Glm-4.5 v and glm-4.1 v-thinking: Towards versatile multimodal reasoning with scalable reinforcement learning.
\newblock \emph{arXiv preprint arXiv:2507.01006}, 2025.

\bibitem[Hu et~al.(2025)Hu, Askari-Hemmat, Hall, Dinan, Zettlemoyer, and Ghazvininejad]{hu2025multimodal}
Yushi Hu, Reyhane Askari-Hemmat, Melissa Hall, Emily Dinan, Luke Zettlemoyer, and Marjan Ghazvininejad.
\newblock Multimodal rewardbench 2: Evaluating omni reward models for interleaved text and image.
\newblock \emph{arXiv preprint arXiv:2512.16899}, 2025.

\bibitem[Hurst et~al.(2024)Hurst, Lerer, Goucher, Perelman, Ramesh, Clark, Ostrow, Welihinda, Hayes, Radford, et~al.]{hurst2024gpt}
Aaron Hurst, Adam Lerer, Adam~P Goucher, Adam Perelman, Aditya Ramesh, Aidan Clark, AJ Ostrow, Akila Welihinda, Alan Hayes, Alec Radford, et~al.
\newblock Gpt-4o system card.
\newblock \emph{arXiv preprint arXiv:2410.21276}, 2024.

\bibitem[Jaech et~al.(2024)Jaech, Kalai, Lerer, Richardson, El-Kishky, Low, Helyar, Madry, Beutel, Carney, et~al.]{jaech2024openai}
Aaron Jaech, Adam Kalai, Adam Lerer, Adam Richardson, Ahmed El-Kishky, Aiden Low, Alec Helyar, Aleksander Madry, Alex Beutel, Alex Carney, et~al.
\newblock Openai o1 system card.
\newblock \emph{arXiv preprint arXiv:2412.16720}, 2024.

\bibitem[Jiang et~al.(2025)Jiang, Zhang, Guo, Li, Qi, Chen, Wang, Jin, Guo, Yan, et~al.]{jiang2025mme}
Dongzhi Jiang, Renrui Zhang, Ziyu Guo, Yanwei Li, Yu Qi, Xinyan Chen, Liuhui Wang, Jianhan Jin, Claire Guo, Shen Yan, et~al.
\newblock Mme-cot: Benchmarking chain-of-thought in large multimodal models for reasoning quality, robustness, and efficiency.
\newblock \emph{arXiv preprint arXiv:2502.09621}, 2025.

\bibitem[Lambert et~al.(2024)Lambert, Pyatkin, Morrison, Miranda, Lin, Chandu, Dziri, Kumar, Zick, Choi, et~al.]{lambert2024rewardbench}
Nathan Lambert, Valentina Pyatkin, Jacob Morrison, LJ Miranda, Bill~Yuchen Lin, Khyathi Chandu, Nouha Dziri, Sachin Kumar, Tom Zick, Yejin Choi, et~al.
\newblock Rewardbench: Evaluating reward models for language modeling.
\newblock \emph{arXiv preprint arXiv:2403.13787}, 2024.

\bibitem[Lee et~al.(2024)Lee, Kim, Park, Kim, and Seo]{lee2024prometheus}
Seongyun Lee, Seungone Kim, Sue Park, Geewook Kim, and Minjoon Seo.
\newblock Prometheus-vision: Vision-language model as a judge for fine-grained evaluation.
\newblock In \emph{Findings of the association for computational linguistics ACL 2024}, pages 11286--11315, 2024.

\bibitem[Li et~al.(2024{\natexlab{a}})Li, Zhang, Guo, Zhang, Li, Zhang, Zhang, Zhang, Li, Liu, et~al.]{li2024llava}
Bo Li, Yuanhan Zhang, Dong Guo, Renrui Zhang, Feng Li, Hao Zhang, Kaichen Zhang, Peiyuan Zhang, Yanwei Li, Ziwei Liu, et~al.
\newblock Llava-onevision: Easy visual task transfer.
\newblock \emph{arXiv preprint arXiv:2408.03326}, 2024{\natexlab{a}}.

\bibitem[Li et~al.(2024{\natexlab{b}})Li, Xie, Li, Chen, Wang, Chen, Yang, Wang, Kong, and Liu]{li2024vlfeedback}
Lei Li, Zhihui Xie, Mukai Li, Shunian Chen, Peiyi Wang, Liang Chen, Yazheng Yang, Benyou Wang, Lingpeng Kong, and Qi Liu.
\newblock Vlfeedback: A large-scale ai feedback dataset for large vision-language models alignment.
\newblock \emph{arXiv preprint arXiv:2410.09421}, 2024{\natexlab{b}}.

\bibitem[Li et~al.(2025{\natexlab{a}})Li, Wei, Xie, Yang, Song, Wang, An, Liu, Li, Lin, et~al.]{li2025vl}
Lei Li, Yuancheng Wei, Zhihui Xie, Xuqing Yang, Yifan Song, Peiyi Wang, Chenxin An, Tianyu Liu, Sujian Li, Bill~Yuchen Lin, et~al.
\newblock Vl-rewardbench: A challenging benchmark for vision-language generative reward models.
\newblock In \emph{Proceedings of the Computer Vision and Pattern Recognition Conference}, pages 24657--24668, 2025{\natexlab{a}}.

\bibitem[Li et~al.(2025{\natexlab{b}})Li, Li, Ren, Tian, Gu, Li, Yue, Wang, Ma, Yang, et~al.]{li2025groundingme}
Rang Li, Lei Li, Shuhuai Ren, Hao Tian, Shuhao Gu, Shicheng Li, Zihao Yue, Yudong Wang, Wenhan Ma, Zhe Yang, et~al.
\newblock Groundingme: Exposing the visual grounding gap in mllms through multi-dimensional evaluation.
\newblock \emph{arXiv preprint arXiv:2512.17495}, 2025{\natexlab{b}}.

\bibitem[Liu et~al.(2023)Liu, Li, Wu, and Lee]{liu2023visual}
Haotian Liu, Chunyuan Li, Qingyang Wu, and Yong~Jae Lee.
\newblock Visual instruction tuning.
\newblock \emph{Advances in neural information processing systems}, 36:\penalty0 34892--34916, 2023.

\bibitem[Liu et~al.(2025)Liu, Liu, Liang, Yuan, Liu, Zheng, Wu, Wang, Xia, Wang, et~al.]{liu2025improving}
Jie Liu, Gongye Liu, Jiajun Liang, Ziyang Yuan, Xiaokun Liu, Mingwu Zheng, Xiele Wu, Qiulin Wang, Menghan Xia, Xintao Wang, et~al.
\newblock Improving video generation with human feedback.
\newblock \emph{arXiv preprint arXiv:2501.13918}, 2025.

\bibitem[Lu et~al.(2024)Lu, Jiang, Chen, Wang, Choi, and Lin]{lu2024wildvision}
Yujie Lu, Dongfu Jiang, Wenhu Chen, William~Yang Wang, Yejin Choi, and Bill~Yuchen Lin.
\newblock Wildvision: Evaluating vision-language models in the wild with human preferences.
\newblock \emph{Advances in Neural Information Processing Systems}, 37:\penalty0 48224--48255, 2024.

\bibitem[Meng et~al.(2025)Meng, Du, Liu, Zhou, Lu, Fu, Shi, Wang, He, Zhang, et~al.]{meng2025mm}
Fanqing Meng, Lingxiao Du, Zongkai Liu, Zhixiang Zhou, Quanfeng Lu, Daocheng Fu, Botian Shi, Wenhai Wang, Junjun He, Kaipeng Zhang, et~al.
\newblock Mm-eureka: Exploring visual aha moment with rule-based large-scale reinforcement learning.
\newblock \emph{arXiv preprint arXiv:2503.07365}, 2025.

\bibitem[Onoe et~al.(2024)Onoe, Rane, Berger, Bitton, Cho, Garg, Ku, Parekh, Pont-Tuset, Tanzer, et~al.]{onoe2024docci}
Yasumasa Onoe, Sunayana Rane, Zachary Berger, Yonatan Bitton, Jaemin Cho, Roopal Garg, Alexander Ku, Zarana Parekh, Jordi Pont-Tuset, Garrett Tanzer, et~al.
\newblock Docci: Descriptions of connected and contrasting images.
\newblock In \emph{European Conference on Computer Vision}, pages 291--309. Springer, 2024.

\bibitem[OpenAI(2025)]{openai2024gpt4_1}
OpenAI.
\newblock Introducing gpt‑4.1 in the api.
\newblock \url{https://openai.com/index/gpt-4-1/}, 2025.

\bibitem[{OpenAI}(2025)]{openai2025o3o4mini}
{OpenAI}.
\newblock Openai o3 and o4-mini system card.
\newblock Webpage, 2025.

\bibitem[PBC(2025)]{anthropic_claude3.7_sonnet_2025}
Anthropic PBC.
\newblock Claude 3.7 sonnet and claude code.
\newblock \url{https://www.anthropic.com/news/claude-3-7-sonnet}, 2025.
\newblock Model release; hybrid reasoning model supporting “extended thinking mode”.

\bibitem[Pitis et~al.(2024)Pitis, Xiao, Le~Roux, and Sordoni]{pitis2024improving}
Silviu Pitis, Ziang Xiao, Nicolas Le~Roux, and Alessandro Sordoni.
\newblock Improving context-aware preference modeling for language models.
\newblock \emph{Advances in Neural Information Processing Systems}, 37:\penalty0 70793--70827, 2024.

\bibitem[Rawal et~al.(2025)Rawal, Shirkavand, Huang, Somepalli, and Goldstein]{rawal2025argus}
Ruchit Rawal, Reza Shirkavand, Heng Huang, Gowthami Somepalli, and Tom Goldstein.
\newblock Argus: Hallucination and omission evaluation in video-llms.
\newblock In \emph{Proceedings of the IEEE/CVF International Conference on Computer Vision}, pages 20280--20290, 2025.

\bibitem[Shirkavand et~al.(2025)Shirkavand, Gao, Yu, and Huang]{shirkavand2025cost}
Reza Shirkavand, Shangqian Gao, Peiran Yu, and Heng Huang.
\newblock Cost-aware contrastive routing for llms.
\newblock \emph{arXiv preprint arXiv:2508.12491}, 2025.

\bibitem[Song et~al.(2025)Song, Ou, Kong, Li, Neubig, and Yue]{song2025visualpuzzles}
Yueqi Song, Tianyue Ou, Yibo Kong, Zecheng Li, Graham Neubig, and Xiang Yue.
\newblock Visualpuzzles: Decoupling multimodal reasoning evaluation from domain knowledge.
\newblock \emph{arXiv preprint arXiv:2504.10342}, 2025.

\bibitem[Sun et~al.(2023)Sun, Shen, Cao, Liu, Li, Shen, Gan, Gui, Wang, Yang, et~al.]{sun2023aligning}
Zhiqing Sun, Sheng Shen, Shengcao Cao, Haotian Liu, Chunyuan Li, Yikang Shen, Chuang Gan, Liang-Yan Gui, Yu-Xiong Wang, Yiming Yang, et~al.
\newblock Aligning large multimodal models with factually augmented rlhf.
\newblock \emph{arXiv preprint arXiv:2309.14525}, 2023.

\bibitem[Tan et~al.(2024)Tan, Zhuang, Montgomery, Tang, Cuadron, Wang, Popa, and Stoica]{tan2024judgebench}
Sijun Tan, Siyuan Zhuang, Kyle Montgomery, William~Y Tang, Alejandro Cuadron, Chenguang Wang, Raluca~Ada Popa, and Ion Stoica.
\newblock Judgebench: A benchmark for evaluating llm-based judges.
\newblock \emph{arXiv preprint arXiv:2410.12784}, 2024.

\bibitem[Team et~al.(2025)Team, Du, Yin, Xing, Qu, Wang, Chen, Zhang, Du, Wei, et~al.]{team2025kimi}
Kimi Team, Angang Du, Bohong Yin, Bowei Xing, Bowen Qu, Bowen Wang, Cheng Chen, Chenlin Zhang, Chenzhuang Du, Chu Wei, et~al.
\newblock Kimi-vl technical report.
\newblock \emph{arXiv preprint arXiv:2504.07491}, 2025.

\bibitem[Tong et~al.(2024)Tong, Brown, Wu, Woo, IYER, Akula, Yang, Yang, Middepogu, Wang, et~al.]{tong2024cambrian}
Peter Tong, Ellis Brown, Penghao Wu, Sanghyun Woo, Adithya Jairam~Vedagiri IYER, Sai~Charitha Akula, Shusheng Yang, Jihan Yang, Manoj Middepogu, Ziteng Wang, et~al.
\newblock Cambrian-1: A fully open, vision-centric exploration of multimodal llms.
\newblock \emph{Advances in Neural Information Processing Systems}, 37:\penalty0 87310--87356, 2024.

\bibitem[Wang et~al.(2025{\natexlab{a}})Wang, Gu, Gao, and Zhou]{wang2025damo}
Kaishen Wang, Hengrui Gu, Meijun Gao, and Kaixiong Zhou.
\newblock Damo: Decoding by accumulating activations momentum for mitigating hallucinations in vision-language models.
\newblock In \emph{The Thirteenth International Conference on Learning Representations}, 2025{\natexlab{a}}.

\bibitem[Wang et~al.(2020)Wang, Wei, Dong, Bao, Yang, and Zhou]{wang2020minilm}
Wenhui Wang, Furu Wei, Li Dong, Hangbo Bao, Nan Yang, and Ming Zhou.
\newblock Minilm: Deep self-attention distillation for task-agnostic compression of pre-trained transformers.
\newblock \emph{Advances in neural information processing systems}, 33:\penalty0 5776--5788, 2020.

\bibitem[Wang et~al.(2025{\natexlab{b}})Wang, Gao, Gu, Pu, Cui, Wei, Liu, Jing, Ye, Shao, et~al.]{wang2025internvl3.5}
Weiyun Wang, Zhangwei Gao, Lixin Gu, Hengjun Pu, Long Cui, Xingguang Wei, Zhaoyang Liu, Linglin Jing, Shenglong Ye, Jie Shao, et~al.
\newblock Internvl3. 5: Advancing open-source multimodal models in versatility, reasoning, and efficiency.
\newblock \emph{arXiv preprint arXiv:2508.18265}, 2025{\natexlab{b}}.

\bibitem[Wang et~al.(2025{\natexlab{c}})Wang, Chen, Wang, Zhou, Zhou, Yao, Zhou, Goldstein, Bhatia, Kass-Hout, et~al.]{wang2025enhancing}
Xiyao Wang, Jiuhai Chen, Zhaoyang Wang, Yuhang Zhou, Yiyang Zhou, Huaxiu Yao, Tianyi Zhou, Tom Goldstein, Parminder Bhatia, Taha Kass-Hout, et~al.
\newblock Enhancing visual-language modality alignment in large vision language models via self-improvement.
\newblock In \emph{Findings of the Association for Computational Linguistics: NAACL 2025}, pages 268--282, 2025{\natexlab{c}}.

\bibitem[Wang et~al.(2025{\natexlab{d}})Wang, Li, Yang, Zhang, Liu, Xiong, and Huang]{wang2025llava}
Xiyao Wang, Chunyuan Li, Jianwei Yang, Kai Zhang, Bo Liu, Tianyi Xiong, and Furong Huang.
\newblock Llava-critic-r1: Your critic model is secretly a strong policy model.
\newblock \emph{arXiv preprint arXiv:2509.00676}, 2025{\natexlab{d}}.

\bibitem[Wang et~al.(2025{\natexlab{e}})Wang, Yang, Feng, Lu, Li, Lin, Lin, Huang, and Wang]{wang2025sota}
Xiyao Wang, Zhengyuan Yang, Chao Feng, Hongjin Lu, Linjie Li, Chung-Ching Lin, Kevin Lin, Furong Huang, and Lijuan Wang.
\newblock Sota with less: Mcts-guided sample selection for data-efficient visual reasoning self-improvement.
\newblock \emph{arXiv preprint arXiv:2504.07934}, 2025{\natexlab{e}}.

\bibitem[Wang et~al.(2025{\natexlab{f}})Wang, Li, Zang, Wang, Lu, Jin, and Wang]{unifiedreward-think}
Yibin Wang, Zhimin Li, Yuhang Zang, Chunyu Wang, Qinglin Lu, Cheng Jin, and Jiaqi Wang.
\newblock Unified multimodal chain-of-thought reward model through reinforcement fine-tuning.
\newblock \emph{arXiv preprint arXiv:2505.03318}, 2025{\natexlab{f}}.

\bibitem[Wang et~al.(2025{\natexlab{g}})Wang, Zang, Li, Jin, and Wang]{wang2025unified}
Yibin Wang, Yuhang Zang, Hao Li, Cheng Jin, and Jiaqi Wang.
\newblock Unified reward model for multimodal understanding and generation.
\newblock \emph{arXiv preprint arXiv:2503.05236}, 2025{\natexlab{g}}.

\bibitem[Wang et~al.(2026)Wang, Zang, Han, Bu, Zhou, Jin, and Wang]{unifiedreward-flex}
Yibin Wang, Yuhang Zang, Feng Han, Jiazi Bu, Yujie Zhou, Cheng Jin, and Jiaqi Wang.
\newblock Unified personalized reward model for vision generation.
\newblock \emph{arXiv preprint arXiv:2602.02380}, 2026.

\bibitem[Wang et~al.(2025{\natexlab{h}})Wang, Zeng, Delalleau, Evans, Egert, Shin, Soares, Dong, and Kuchaiev]{wang2025rlbff}
Zhilin Wang, Jiaqi Zeng, Olivier Delalleau, Ellie Evans, Daniel Egert, Hoo-Chang Shin, Felipe Soares, Yi Dong, and Oleksii Kuchaiev.
\newblock Rlbff: Binary flexible feedback to bridge between human feedback \& verifiable rewards.
\newblock \emph{arXiv preprint arXiv:2509.21319}, 2025{\natexlab{h}}.

\bibitem[Xie et~al.(2025)Xie, Kong, Dong, Sima, Zhang, Chen, Liu, and Pan]{xie2025vlms}
Shaoyuan Xie, Lingdong Kong, Yuhao Dong, Chonghao Sima, Wenwei Zhang, Qi~Alfred Chen, Ziwei Liu, and Liang Pan.
\newblock Are vlms ready for autonomous driving? an empirical study from the reliability, data, and metric perspectives.
\newblock \emph{arXiv preprint arXiv:2501.04003}, 2025.

\bibitem[Xiong et~al.(2024)Xiong, Li, Guo, Yuan, Gu, and Li]{xiong2024llavaovchat}
Tianyi Xiong, Bo Li, Dong Guo, Huizhuo Yuan, Quanquan Gu, and Chunyuan Li.
\newblock Llava-onevisionchat: Improving chat with preference learning, 2024.

\bibitem[Xiong et~al.(2025)Xiong, Wang, Guo, Ye, Fan, Gu, Huang, and Li]{xiong2025llava}
Tianyi Xiong, Xiyao Wang, Dong Guo, Qinghao Ye, Haoqi Fan, Quanquan Gu, Heng Huang, and Chunyuan Li.
\newblock Llava-critic: Learning to evaluate multimodal models.
\newblock In \emph{Proceedings of the Computer Vision and Pattern Recognition Conference}, pages 13618--13628, 2025.

\bibitem[Xiong et~al.(2026)Xiong, Wang, Liu, Dong, Li, Huang, Kautz, and Yu]{xiong2026phycritic}
Tianyi Xiong, Shihao Wang, Guilin Liu, Yi Dong, Ming Li, Heng Huang, Jan Kautz, and Zhiding Yu.
\newblock Phycritic: Multimodal critic models for physical ai.
\newblock \emph{arXiv preprint arXiv:2602.11124}, 2026.

\bibitem[Yang et~al.(2025)Yang, Li, Yang, Zhang, Hui, Zheng, Yu, Gao, Huang, Lv, et~al.]{yang2025qwen3}
An Yang, Anfeng Li, Baosong Yang, Beichen Zhang, Binyuan Hui, Bo Zheng, Bowen Yu, Chang Gao, Chengen Huang, Chenxu Lv, et~al.
\newblock Qwen3 technical report.
\newblock \emph{arXiv preprint arXiv:2505.09388}, 2025.

\bibitem[Yasunaga et~al.(2025)Yasunaga, Zettlemoyer, and Ghazvininejad]{yasunaga2025multimodal}
Michihiro Yasunaga, Luke Zettlemoyer, and Marjan Ghazvininejad.
\newblock Multimodal rewardbench: Holistic evaluation of reward models for vision language models.
\newblock \emph{arXiv preprint arXiv:2502.14191}, 2025.

\bibitem[Ye et~al.(2025)Ye, Zeng, Li, Li, and Fan]{ye2025painting}
Qinghao Ye, Xianhan Zeng, Fu Li, Chunyuan Li, and Haoqi Fan.
\newblock Painting with words: Elevating detailed image captioning with benchmark and alignment learning.
\newblock \emph{arXiv preprint arXiv:2503.07906}, 2025.

\bibitem[Yu et~al.(2016)Yu, Poirson, Yang, Berg, and Berg]{yu2016modeling}
Licheng Yu, Patrick Poirson, Shan Yang, Alexander~C Berg, and Tamara~L Berg.
\newblock Modeling context in referring expressions.
\newblock In \emph{European conference on computer vision}, pages 69--85. Springer, 2016.

\bibitem[Yu et~al.(2025{\natexlab{a}})Yu, Wang, Wang, Huang, Ma, He, Cai, Chen, Huang, Zhao, et~al.]{yu2025minicpm}
Tianyu Yu, Zefan Wang, Chongyi Wang, Fuwei Huang, Wenshuo Ma, Zhihui He, Tianchi Cai, Weize Chen, Yuxiang Huang, Yuanqian Zhao, et~al.
\newblock Minicpm-v 4.5: Cooking efficient mllms via architecture, data, and training recipe.
\newblock \emph{arXiv preprint arXiv:2509.18154}, 2025{\natexlab{a}}.

\bibitem[Yu et~al.(2025{\natexlab{b}})Yu, Zhang, Li, Xu, Yao, Chen, Lu, Cui, Dang, He, et~al.]{yu2025rlaif}
Tianyu Yu, Haoye Zhang, Qiming Li, Qixin Xu, Yuan Yao, Da Chen, Xiaoman Lu, Ganqu Cui, Yunkai Dang, Taiwen He, et~al.
\newblock Rlaif-v: Open-source ai feedback leads to super gpt-4v trustworthiness.
\newblock In \emph{Proceedings of the Computer Vision and Pattern Recognition Conference}, pages 19985--19995, 2025{\natexlab{b}}.

\bibitem[Yu et~al.(2023)Yu, Yang, Li, Wang, Lin, Liu, Wang, and Wang]{yu2023mm}
Weihao Yu, Zhengyuan Yang, Linjie Li, Jianfeng Wang, Kevin Lin, Zicheng Liu, Xinchao Wang, and Lijuan Wang.
\newblock Mm-vet: Evaluating large multimodal models for integrated capabilities.
\newblock \emph{arXiv preprint arXiv:2308.02490}, 2023.

\bibitem[Zang et~al.(2025)Zang, Dong, Zhang, Cao, Liu, Ding, Wu, Ma, Duan, Zhang, et~al.]{zang2025internlm}
Yuhang Zang, Xiaoyi Dong, Pan Zhang, Yuhang Cao, Ziyu Liu, Shengyuan Ding, Shenxi Wu, Yubo Ma, Haodong Duan, Wenwei Zhang, et~al.
\newblock Internlm-xcomposer2. 5-reward: A simple yet effective multi-modal reward model.
\newblock \emph{arXiv preprint arXiv:2501.12368}, 2025.

\bibitem[Zeng et~al.(2025)Zeng, Luo, Lin, Tian, Li, Gong, Guo, and Ma]{zeng-etal-2025-mm}
Gailun Zeng, Ziyang Luo, Hongzhan Lin, Yuchen Tian, Kaixin Li, Ziyang Gong, Jianxiong Guo, and Jing Ma.
\newblock {MM}-{CRITIC}: A holistic evaluation of large multimodal models as multimodal critique.
\newblock In \emph{Findings of the Association for Computational Linguistics: EMNLP 2025}, pages 13603--13630, Suzhou, China, 2025. Association for Computational Linguistics.

\bibitem[Zhang et~al.(2024)Zhang, Jiang, Zhang, Lin, Guo, Qiu, Zhou, Lu, Chang, Qiao, et~al.]{zhang2024mathverse}
Renrui Zhang, Dongzhi Jiang, Yichi Zhang, Haokun Lin, Ziyu Guo, Pengshuo Qiu, Aojun Zhou, Pan Lu, Kai-Wei Chang, Yu Qiao, et~al.
\newblock Mathverse: Does your multi-modal llm truly see the diagrams in visual math problems?
\newblock In \emph{European Conference on Computer Vision}, pages 169--186. Springer, 2024.

\bibitem[Zhang et~al.(2023)Zhang, Lu, Wang, Yan, Yan, Qin, Wang, Yan, Wang, and Petzold]{zhang2023gpt}
Xinlu Zhang, Yujie Lu, Weizhi Wang, An Yan, Jun Yan, Lianke Qin, Heng Wang, Xifeng Yan, William~Yang Wang, and Linda~Ruth Petzold.
\newblock Gpt-4v (ision) as a generalist evaluator for vision-language tasks.
\newblock \emph{arXiv preprint arXiv:2311.01361}, 2023.

\bibitem[Zhang et~al.(2025{\natexlab{a}})Zhang, Lu, Hu, Fu, Wen, Zhang, Liu, Jiang, Chen, Tang, et~al.]{zhang2025r1}
Yi-Fan Zhang, Xingyu Lu, Xiao Hu, Chaoyou Fu, Bin Wen, Tianke Zhang, Changyi Liu, Kaiyu Jiang, Kaibing Chen, Kaiyu Tang, et~al.
\newblock R1-reward: Training multimodal reward model through stable reinforcement learning.
\newblock \emph{arXiv preprint arXiv:2505.02835}, 2025{\natexlab{a}}.

\bibitem[Zhang et~al.(2025{\natexlab{b}})Zhang, Yang, Zhang, Shi, Chen, Tian, Fu, Wang, Wu, Cui, et~al.]{zhang2025basereward}
Yi-Fan Zhang, Haihua Yang, Huanyu Zhang, Yang Shi, Zezhou Chen, Haochen Tian, Chaoyou Fu, Haotian Wang, Kai Wu, Bo Cui, et~al.
\newblock Basereward: A strong baseline for multimodal reward model.
\newblock \emph{arXiv preprint arXiv:2509.16127}, 2025{\natexlab{b}}.

\bibitem[Zhang et~al.(2025{\natexlab{c}})Zhang, Yu, Tian, Fu, Li, Zeng, Xie, Shi, Zhang, Wu, et~al.]{zhang2025mm}
Yi-Fan Zhang, Tao Yu, Haochen Tian, Chaoyou Fu, Peiyan Li, Jianshu Zeng, Wulin Xie, Yang Shi, Huanyu Zhang, Junkang Wu, et~al.
\newblock Mm-rlhf: The next step forward in multimodal llm alignment.
\newblock \emph{arXiv preprint arXiv:2502.10391}, 2025{\natexlab{c}}.

\bibitem[Zheng et~al.(2025{\natexlab{a}})Zheng, Chen, Han, McCoy, and Huang]{zheng2025learning}
Tong Zheng, Lichang Chen, Simeng Han, R~Thomas McCoy, and Heng Huang.
\newblock Learning to reason via mixture-of-thought for logical reasoning.
\newblock \emph{arXiv preprint arXiv:2505.15817}, 2025{\natexlab{a}}.

\bibitem[Zheng et~al.(2025{\natexlab{b}})Zheng, Zhang, Yu, Wang, Dai, Liu, Bao, Huang, Huang, and Yu]{zheng2025parallel}
Tong Zheng, Hongming Zhang, Wenhao Yu, Xiaoyang Wang, Runpeng Dai, Rui Liu, Huiwen Bao, Chengsong Huang, Heng Huang, and Dong Yu.
\newblock Parallel-r1: Towards parallel thinking via reinforcement learning.
\newblock \emph{arXiv preprint arXiv:2509.07980}, 2025{\natexlab{b}}.

\bibitem[Zheng et~al.(2026)Zheng, Huang, Dai, He, Liu, Ni, Bao, Wang, Zhu, Huang, et~al.]{zheng2026parallel}
Tong Zheng, Chengsong Huang, Runpeng Dai, Yun He, Rui Liu, Xin Ni, Huiwen Bao, Kaishen Wang, Hongtu Zhu, Jiaxin Huang, et~al.
\newblock Parallel-probe: Towards efficient parallel thinking via 2d probing.
\newblock \emph{arXiv preprint arXiv:2602.03845}, 2026.

\bibitem[Zhou et~al.(2024)Zhou, Zheng, Wang, Xi, Dou, Bao, Shen, Xiong, Fan, Mou, et~al.]{zhou2024rmb}
Enyu Zhou, Guodong Zheng, Binghai Wang, Zhiheng Xi, Shihan Dou, Rong Bao, Wei Shen, Limao Xiong, Jessica Fan, Yurong Mou, et~al.
\newblock Rmb: Comprehensively benchmarking reward models in llm alignment.
\newblock \emph{arXiv preprint arXiv:2410.09893}, 2024.

\bibitem[Zhu et~al.(2025)Zhu, Wang, Chen, Liu, Ye, Gu, Tian, Duan, Su, Shao, et~al.]{zhu2025internvl3}
Jinguo Zhu, Weiyun Wang, Zhe Chen, Zhaoyang Liu, Shenglong Ye, Lixin Gu, Hao Tian, Yuchen Duan, Weijie Su, Jie Shao, et~al.
\newblock Internvl3: Exploring advanced training and test-time recipes for open-source multimodal models.
\newblock \emph{arXiv preprint arXiv:2504.10479}, 2025.

\end{thebibliography}
}

\clearpage
\appendix

\section{Overview of the Appendix}

\begin{itemize}[itemsep=3pt, topsep=3pt]

    \item Section~\ref{sec:appendix-comparison-benchmarks} illustrates the differences between \ours and prior multimodal judge benchmarks.
    
    \item Section~\ref{sec:appendix-benchmark-details} provides additional benchmark details, including the response models, evaluation criteria, judge prompt template, and supplementary data statistics.

    \item Section~\ref{sec:appendix-data-example} presents qualitative examples of multimodal prompts and complete evaluation instances.

    \item Section~\ref{sec:appendix-experients} reports additional results and analysis.
\end{itemize}

\section{Comparison Against Prior Benchmarks}
\label{sec:appendix-comparison-benchmarks}

As shown in Table~\ref{tab:appendix-compare-benchmarks}, earlier benchmarks assign only a single overall preference label for each pair of model responses, whereas \ours provides multiple criterion-level human annotations. This makes \ours the first benchmark capable of assessing whether LMM judges can follow pluralistic evaluation criteria and deliver reliable criterion-level judgments.
In addition, the average length of candidate model responses in \ours is significantly longer (164 vs.\ 99 words), further highlighting the increased difficulty and overall challenge of the benchmark.

\definecolor{darkgreen}{RGB}{0,138,0}
\begin{table}[!h]
    \centering
    \vspace{-1.5mm}
    \setlength{\tabcolsep}{2pt}
    \small
    \scalebox{0.88}{
    \begin{tabular}{c|cc}
    \toprule
       Benchmark  & Resp. Quantile & Multi-Criterion Anno. \\
    \midrule
       MLLM-as-a-Judge~\cite{chen2024mllm}  & (54, 89, 153) & \textcolor{red}{\xmark} \\
       VL-RewardBench~\cite{li2025vl} & (48, 99, 136)&  \textcolor{red}{\xmark} \\
       MM-RLHF~\cite{zhang2025mm} & (16, 49, 144) &  \textcolor{red}{\xmark} \\
       Multimodal-RewardBench~\cite{yasunaga2025multimodal} & (1, 72, 138) & \textcolor{red}{\xmark}  \\
    \hline
        \ours (ours) & (104, 164, 247) & \textcolor{darkgreen}{\cmark}\\   
    \bottomrule
    \end{tabular}
    }
    \vspace{-1.0mm}
    \caption{
    Comparison with existing benchmarks.
    }
    \vspace{-4mm}
    \label{tab:appendix-compare-benchmarks}
\end{table}

\section{Benchmark Construction and Statistics}

\label{sec:appendix-benchmark-details}

\subsection{Additional Benchmark Details}

\noindent\textbf{Models for Generating Candidate Responses:} Gemini-2.5-Pro~\cite{comanici2025gemini}, Gemini-2.5-Flash~\cite{comanici2025gemini}, GPT-4.1~\cite{openai2024gpt4_1}, GPT-4o~\cite{hurst2024gpt}, Qwen2.5-VL-7B-Instruct~\cite{bai2025qwen2}, Qwen2.5-VL-72B-Instruct~\cite{bai2025qwen2}, LLaMA-3.2-11B-Vision-Instruct~\cite{grattafiori2024llama}, InternVL3-8B-Instruct~\cite{zhu2025internvl3}, InternVL3-38B-Instruct~\cite{zhu2025internvl3}, MiMo-VL-7B-RL~\cite{team2025kimi}, GLM-4.1V-9B-Thinking~\cite{hong2025glm}.

Specifically, for prompts from WildVision-Battle~\cite{lu2024wildvision}, if the two original responses receive a tie in their human preference, we keep the original pair; otherwise, we generate new responses using the models listed above.

\vspace{2mm}
\noindent\textbf{Detailed Evaluation Criteria.}
Table~\ref{tab:appendix_criteria_open} presents the detailed criteria for judging open-ended generation, and Table~\ref{tab:appendix_criteria_reasoning} lists those for judging verifiable reasoning.
To better guide both human annotators and judge LMMs, we organize the description of each criterion into three components:
\begin{enumerate}[itemsep=3pt, topsep=3pt]
    \item \textbf{Definition}: A general descriptions of the criterion and the key aspects a response should demonstrate.
    \item \textbf{Positive Indicators}: Behaviors and qualities that are encouraged and should be preferred in responses.
    \item \textbf{Negative Indicators}: Errors, undesirable behaviors, or deficiencies that should be penalized under this criterion.
\end{enumerate}

\vspace{2mm}
\noindent\textbf{Evaluation Prompt Template.}
We provide the evaluation prompt used for judge-model inference in Table~\ref{tab:append_prompt_template}. 
When evaluating LMM judges on the \ours benchmark, each criterion instance—consisting of the question, the pair of responses, and the target criterion—is filled into this template and sent to the model during inference. 
For finetuned judges~\cite{xiong2025llava,wang2025llava,zhang2025r1,wang2025unified}, we adapt this template to match the prompt format used in their original repositories, avoiding significant out-of-distribution judgment behaviors.

\subsection{Additional Data Statistics}

\noindent\textbf{Response Pair Composition.}
As shown in Table~\ref{tab:appendix-response-pair-types}, for both splits, roughly $40\%$ of response pairs originate from the same model via random sampling, while the remaining pairs are generated by different models. 
These two types serve complementary roles:
cross-model pairs reflect realistic deployment scenarios for automated evaluation (e.g., LMArena or WildVision Arena), where responses differ along multiple dimensions;
intra-model pairs control for model identity and surface style, which is crucial when applying judges for test-time scaling and RL reward modeling.
Our balanced mix ensures that \ours assesses whether LMM judges can capture quality differences both across models (reflecting model diversity) and within the same model (reflecting sampling diversity).

\begin{table}[!h]
    \centering
    \small
    \vspace{-1mm}
    \setlength{\tabcolsep}{3pt}
    \begin{tabular}{lccc}
        \toprule
        Split & \#Prompts & Cross-model pairs & Intra-model pairs \\
        \midrule
        Open-ended   & 299 & 168 \;(56.2\%) & 131 \;(43.8\%) \\
        Reasoning    & 126 &  76 \;(60.3\%) &  50 \;(39.7\%) \\
        \midrule
        Total        & 425 & 244 \;(57.4\%) & 181 \;(42.6\%) \\
        \bottomrule
    \end{tabular}
    \vspace{-2mm}
    \caption{Composition of response pairs.}
    \vspace{-3mm}
    \label{tab:appendix-response-pair-types}
\end{table}

\begin{figure*}
    \centering
    \includegraphics[width=0.85\linewidth]{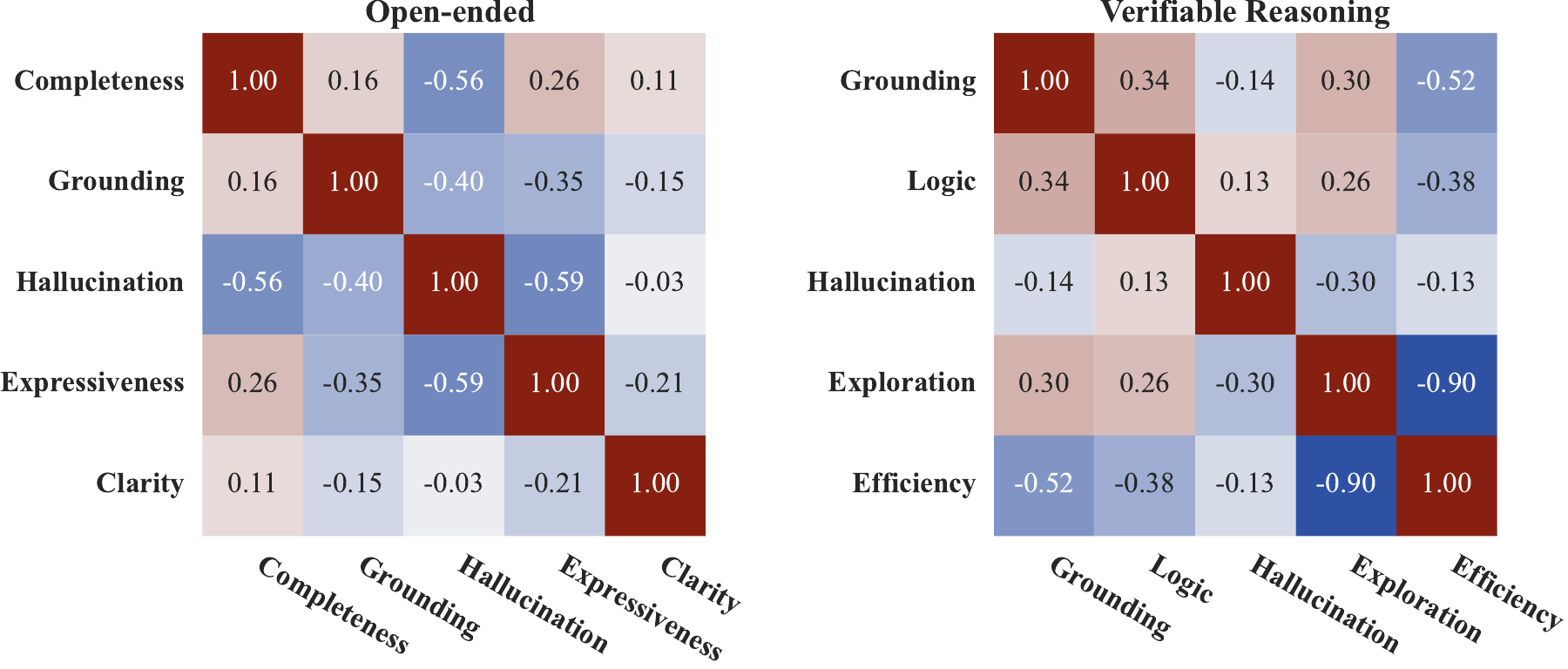}
    \vspace{-2.5mm}
    \caption{Correlation of criterion-level human preferences for prompts that exhibit preference conflicts in the open-ended (\textit{left}) and reasoning (\textit{right}) splits of \ours.}
    
    \label{fig:appendix-criterion-correlation-prompt-conflict}
\end{figure*}

\begin{figure*}
    \centering
    \includegraphics[width=0.85\linewidth]{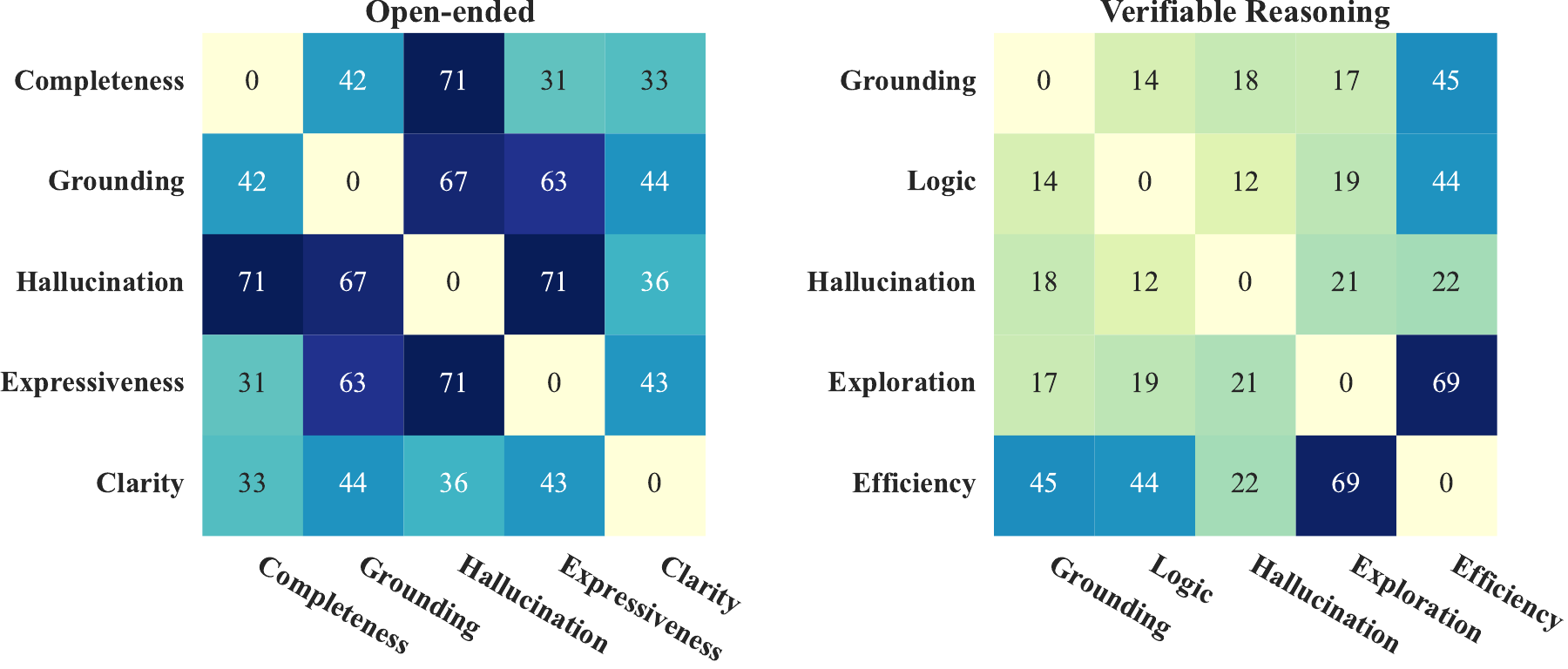}
    \vspace{-2.5mm}
    \caption{Counts of criterion pairs exhibiting human preference conflicts in the open-ended (\textit{left}) and reasoning splits (\textit{right}) of \ours.}
    \label{fig:criterion-inconsistent-counts}
\end{figure*}

\vspace{2mm}
\noindent\textbf{Cross-Criterion Correlation and Conflict Frequency.}
We report additional statistics on criterion-level human preferences to illustrate how \ours captures cross-criterion correlations and conflicts.
Figure~\ref{fig:appendix-criterion-correlation-prompt-conflict} shows how human preferences correlate across criteria on prompts that exhibit criterion-level conflicts. Figure~\ref{fig:criterion-inconsistent-counts} reports how often two criteria yield opposite human preferences on the same response pair.
\ours contains preference conflicts across all criterion pairs in both splits, reflecting both the nuanced differences within the response pairs and the inherent diversity and tension across evaluation dimensions. 
Certain criterion pairs display intuitive trade-offs—such as \textit{completeness vs. no-hallucination} and \textit{no-hallucination vs. expressiveness} in the open-ended split, and \textit{exploration vs. efficiency} in the reasoning split—and these pairs also exhibit the highest frequencies of human-annotated conflicts.

\begin{table*}[!tp]
\begin{minipage}{1.0\textwidth}
\centering
\scalebox{1}{
    \begin{tabular}{l p{11.5cm}}
    \toprule
    \textbf{Evaluation Criterion} & \textbf{Descrption} \\
    \midrule
    \textbf{Completeness and Coverage} & The response should provide a thorough and well-developed answer that fully addresses the intent of the prompt. It must address the complete scope of the task, incorporating all major elements specified in the prompt, as well as relevant visual aspects and broader contextual cues. When appropriate, drawing on relevant external knowledge is encouraged to enrich the explanation.  \\
    & - Reward: The response addresses all key parts of the prompt and image, showing depth and effort in the description or explanation.  \\
    & - Penalize: The response is underdeveloped, fails to meet one or more specific requirements in the prompt, or omits important visual elements or interpretive points.\\
    \midrule

   \textbf{Visual Grounding and Details} &  The response should demonstrate a clear and meaningful connection to the visual input. It should refer to observable elements in the image—such as objects, spatial relationships, colors, or text—and build its description or interpretation based on those elements.  \\
    & - Reward: The response explicitly references relevant visual details that are clearly visible in the image.  \\
    & - Penalize: The response fails to connect meaningfully to the image, or uses vague, generic language that lacks specific visual grounding. \\

    \midrule

    \textbf{Factuality / No Hallucination} &  The response should avoid introducing any visual details, objects, relationships, or factual claims that are not present in the image or reasonably suggested by the prompt. This includes both visual hallucinations (e.g., describing elements not visible in the image) and factual inaccuracies in general knowledge.  \\
    & - Reward: The response stays grounded in the image and prompt, without inventing visual elements or making unsupported factual claims.  \\
    & - Penalize: The response introduces hallucinated visual content or inaccurate factual statements that are unsupported or misleading. \\

    \midrule

   \textbf{Creativity and Expressiveness} &  The response should show originality or stylistic flair for open-ended tasks, and knowledge-informed articulation with precision and depth for analytical tasks. All responses must remain contextually appropriate and grounded in the visual input, while enhancing richness, nuance, and overall engagement. \\
    & - Reward: The response uses vivid language, unique phrasing, or inventive associations that enrich the interpretation, or it demonstrates professional articulation through deep and knowledge-grounded analysis. \\
    & - Penalize: The response is overly literal, flat, or dull, lacking originality, variation in expression, or in analytical contexts, fails to demonstrate professional depth or expertise. \\

    \midrule

   \textbf{Clarity and Coherence}  & The response should communicate ideas clearly and logically, with coherent structure and fluent language. This involves not only grammatical correctness, but also effective organization of information, smooth transitions, and consistent flow of ideas.  \\
    & - Reward: The response is clearly written, logically structured, and easy to follow. A brief summary at the beginning may further improve clarity.  \\
    & - Penalize: The response is difficult to follow due to unclear structure, disorganized reasoning, poor transitions, or awkward and repetitive phrasing. \\
    \bottomrule
    \end{tabular}
}
\captionof{table}{Five evaluation criteria for judging \textbf{\textit{open-ended generation}} tasks in \ours Benchmark.}
\label{tab:appendix_criteria_open}  
\end{minipage}
\end{table*}

\begin{table*}[!tp]
\begin{minipage}{0.95\textwidth}
\centering
\scalebox{1}{
    \begin{tabular}{p{4cm} p{11.5cm}}
    \toprule
    \textbf{Evaluation Criterion} & \textbf{Descrption} \\
    \midrule
    \textbf{Visual Grounding}  & The response should be explicitly grounded in the visual input. It must refer to salient visual elements—such as specific objects, spatial arrangements, colors, or visible text—and incorporate them meaningfully into the reasoning process. Visual references should be accurate and relevant to the task.  \\
    & - Reward: The response clearly references important visual features and integrates them into the reasoning in a precise and relevant manner. \\ 
    & - Penalize: The response fails to reference relevant visual elements, or uses generic or weakly connected visual details that do not meaningfully support the reasoning. \\

    \midrule

    \textbf{Logic Coherence and Consistency} & The reasoning should follow a logically sound and step-by-step progression, with each step building upon the previous one. The reasoning should be internally consistent, with no contradictions, missing steps, or unjustified leaps. The final answer should naturally and justifiably emerge from the reasoning process.  \\
    & - Reward: The response presents a well-structured, internally consistent chain of reasoning that leads clearly and justifiably to the final answer.  \\
    & - Penalize: The response contains contradictions, missing steps, or disconnects between reasoning and answer. Short-cut behaviors—such as giving the final answer first with unsupported or inconsistent reasoning—should also be penalized. \\

    \midrule

    \textbf{Factuality / No Hallucination} & All claims and reasoning steps must be factually accurate and supported by the image or the prompt. The response should avoid hallucinated visual content, misidentifications, or factual inaccuracies in the reasoning process.  \\
    & - Reward: The response is free from factual errors or hallucinations and relies only on valid observations and logical inferences.  \\
    & - Penalize: The response introduces hallucinated details, misidentifications, or incorrect factual claims that compromise the reasoning. \\

    \midrule

    \textbf{Reflection and Exploration} & The reasoning should demonstrate thoughtful reflection and a willingness to explore multiple possibilities, particularly when the task is ambiguous or complex. This includes acknowledging uncertainty, considering alternative interpretations, or revising initial assumptions.  \\
    & - Reward: The response demonstrates depth through reflection, critical evaluation, or exploration of different solutions before reaching a conclusion.  \\
    & - Penalize: The response is overly rigid, superficial, or rushed, showing little to no depth of thought, reflection, or exploration of alternative possibilities. \\

    \midrule
    
    \textbf{Conciseness and Efficiency} & The reasoning should be clear, focused, and efficiently communicate the steps. It should avoid redundancy, digressions, or unnecessary elaboration that dilute the argument. For straightforward tasks, over-explaining or over-analyzing should also be avoided. \\  
& - Reward: The response is concise and well-structured, conveying reasoning steps precisely and proportionally to the task complexity.  \\
& - Penalize: The response is verbose, repetitive, or includes irrelevant content that distracts from the reasoning. It may also overthink or over-explain simple prompts. \\
    \bottomrule
    \end{tabular}
}
\captionof{table}{Five evaluation criteria for judging \textbf{\textit{verifiable reasoning}} tasks in \ours benchmark.}
\label{tab:appendix_criteria_reasoning}  
\end{minipage}
\end{table*}

\begin{table*}[!htp]
\centering
\begin{minipage}{1.0\textwidth}
\vspace{0mm}    
\centering
\begin{tcolorbox} 
\centering
\begin{tabular}{p{0.96\textwidth}}
You are an expert in evaluating the quality of AI-generated responses according to specific evaluation criteria. Your task is to assess two responses generated by different AI assistants in reply to a user’s question about an image. The image is provided as part of the input. 

You must evaluate the responses **strictly and exclusively** based on the following evaluation criterion:

\texttt{\{Criterion\}}

Do not consider any other dimensions or criteria beyond what is specified above. 

Here are the inputs for your evaluation:

[Question]: \texttt{\{Question\}} \\

[Response 1]: \texttt{\{Response1\}} \\

[Response 2]: \texttt{\{Response2\}} \\

First, provide a detailed justification for your evaluation. Refer to specific elements in the responses, how they align with the evaluation criterion, and relevant visual details from the image. \\

On the final line, provide your final judgment on which response is better. Your judgment must be based solely on the specified criterion. Strictly follow this format:  
Response X is better.

\end{tabular}
\end{tcolorbox}
\vspace{-2.5mm}
\caption{Evaluation prompt template used for LMM judge inference in \ours. Judge models are explicitly instructed to focus strictly and exclusively on the target evaluation criterion during each inference. }
\label{tab:append_prompt_template}
\end{minipage}
\end{table*}

\clearpage
\clearpage

\section{Qualitative Examples}
\label{sec:appendix-data-example}

\subsection{Multimodal Prompts}

\begin{figure*}[!htp]
    \centering
    \vspace{-2mm}
    \includegraphics[width=0.93\linewidth]{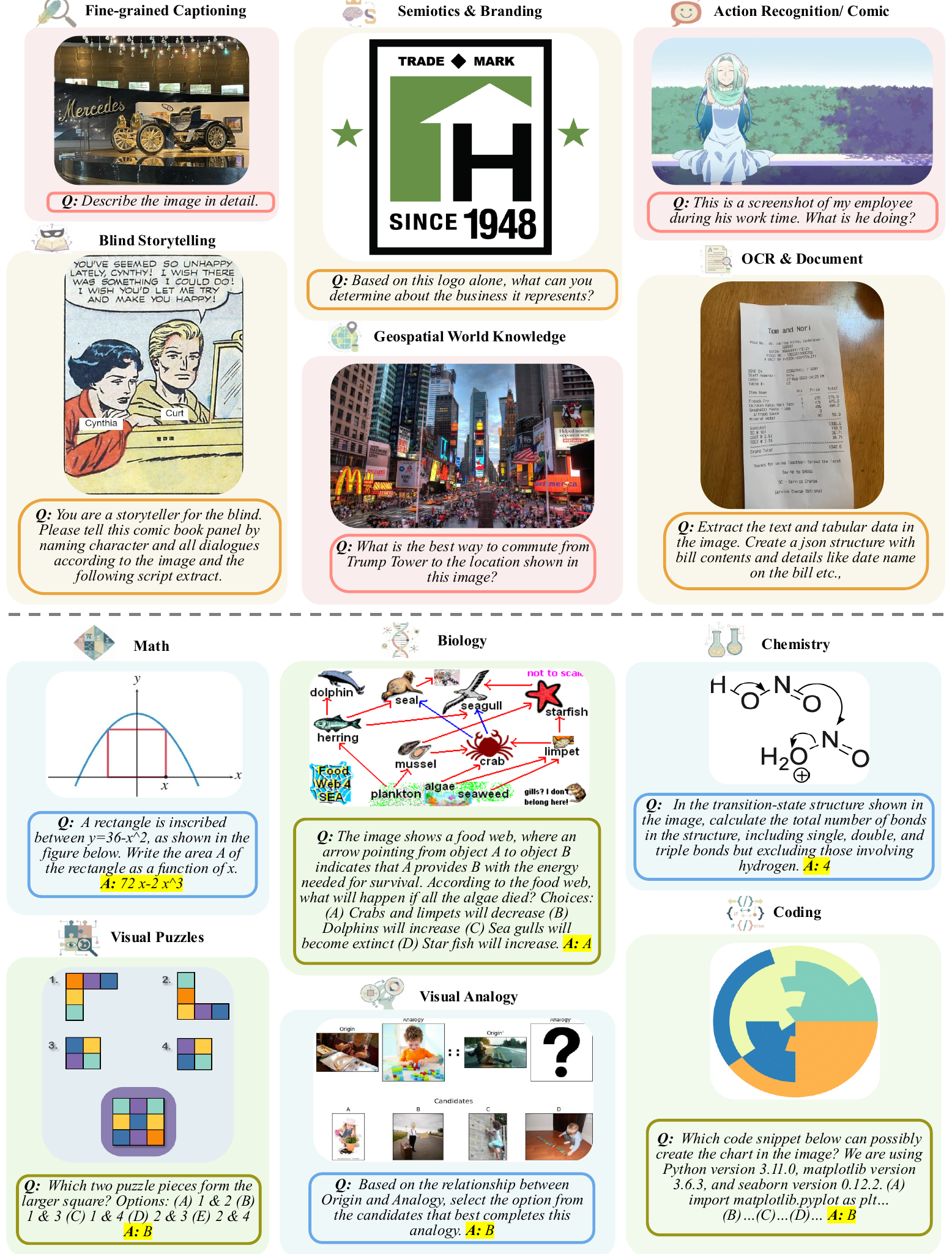}

    \caption{
        \textbf{Example prompts in the open-ended (\textit{top}) and verifiable reasoning (\textit{bottom}) splits of \ours.}
        Our benchmark spans diverse multimodal prompts, reflecting the broad evaluation domains in which LMM judges are used.
    }
    
    \label{fig:appendix-prompt-example}
\end{figure*}

Figure~\ref{fig:appendix-prompt-example} visualizes selected multimodal prompts from \ours, spanning diverse \textit{open-ended generation} scenarios—such as captioning, storytelling, knowledge-based tasks, and text-rich understanding—as well as \textit{verifiable reasoning} domains including math, science, coding, and visual analogy. These prompts cover a wide range of LMM-as-a-Judge use cases.

\subsection{Full Evaluation Instances}

We present six full evaluation instances—each consisting of a prompt and paired model responses—along with their multi-criterion human preference annotations. These examples illustrate how our curated data reflects fine-grained response differences and demonstrates criterion-level preference conflict patterns within each response pair.

The first three examples illustrate judgment in open-ended tasks:

\begin{itemize}
[itemsep=2pt,topsep=2pt]
    \item Table~\ref{tab:example_openended_conflict_blind} shows an evaluation instance for judging two models in blind storytelling. This example reveals preference conflicts such as \textit{completeness vs.\ grounding} and \textit{grounding vs.\ no-hallucination}.
    
    \item Table~\ref{tab:example_openended_conflict_beerpool} presents an evaluation instance for judging two models on captioning creative images. It exhibits representative preference conflicts including \textit{completeness vs.\ no-hallucination}, \textit{completeness vs.\ expressiveness}, and  \textit{expressiveness vs.\ clarity}.
    
    \item Table~\ref{tab:example_openended_conflict_kids} provides an example of judging two responses from the same model on an instruction-rich image analysis task. Observed conflicts include \textit{grounding vs.\ clarity} and the less common \textit{no-hallucination vs.\ clarity}.
\end{itemize}

The following three examples illustrate judgment in verifiable reasoning tasks:

\begin{itemize} [itemsep=2pt,topsep=2pt]
    \item Table~\ref{tab:verifiable_reasoning_symmetry} shows an evaluation instance for judging two models' thinking traces on a math reasoning task. Although both traces lead to incorrect answers, one exhibits better \textit{visual grounding} and \textit{explores} alternative answers, whereas the other maintains stronger \textit{logical consistency} with \textit{fewer perceptual hallucinations}. 
    
    \item Table~\ref{tab:verifiable_reasoning_buses} presents an evaluation instance for judging two thinking traces produced by the same model on object counting. While both responses lead to the correct final answer, one conducts \textit{concise} reasoning, whereas the other identifies more \textit{visual details} and actively \textit{reflects on each object}, but introduces \textit{minor hallucinations} that do not affect the final result.
    
    \item Table~\ref{tab:verifiable_reasoning_foodweb} provides an example of judging two models’ responses on a biological reasoning task involving a food web. In this case, one response is more \textit{concise} and demonstrates deeper \textit{reflection} by explicitly identifying a key intermediate inference, while the other offers a more visually detailed description of the food-web structure.
\end{itemize}

\setul{0.6ex}{0.8pt}
\newcommand{\coloreduline}[2]{%
  {\setulcolor{#1}\ul{#2}}%
}
\renewcommand{\ULthickness}{1pt} 

%

\begin{table*}[!ht]
\begin{minipage}{1.0\textwidth}
\centering
\scalebox{0.98}{
\renewcommand{\arraystretch}{1.15}
\setlength{\tabcolsep}{3pt}
\begin{tabular}{l p{13cm}}
\rowcolor{tbhead}
\toprule
\multicolumn{2}{l}{\bfseries Open-ended Example 1} \\
\midrule
& \includegraphics[width=3.8cm]{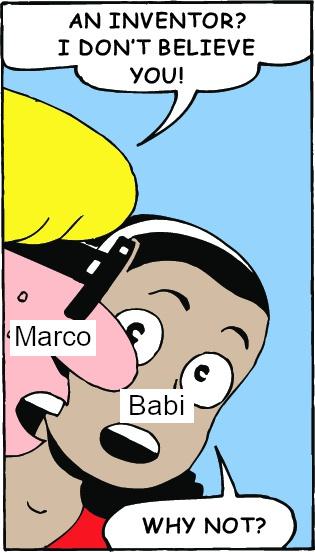} \\[0.3em]
Question &
You are a storyteller for the blind. Please tell this comic book panel by naming character according to the image and the following script extract. Corresponding character’s names have been added in white rectangles over the original image (\textcolor{critTextA}{do not mention rectangle or label in the storytelling}).
\\[0.4em]
& DIALOGUE \\[-0.2em]
& Marco: AN INVENTOR? I DON'T BELIEVE YOU! \\
& Babi: WHY NOT? \\

\midrule
\arrayrulecolor{grid}

\rowcolor{tbshade}
\makecell[lt]{Response A \\ (llava-v1.6-vicuna-7b)} &
In the comic book panel, a character is named Marco. He is speaking to another character named Babi. Marco is questioning Babi's claim that he is an inventor. Babi responds with an open-ended question, encouraging Marco to reconsider his skepticism. Marco's skepticism towards Babi's claim is evident from his words, ``I DON'T BELIEVE YOU!'' It seems like Marco is questioning Babi's credibility or the authenticity of his claim. Babi's open-ended question, ``WHY NOT?'', indicates that he is not backing down and is pushing for a dialogue or a conversation where Marco can clarify his doubts. This comic panel seems to be a \textcolor{critTextA}{humorous take on skepticism and the importance of asking questions to explore different perspectives and possibilities}.
\\
\midrule
\makecell[lt]{Response B \\ (yi-vl-plus)} &
In a vividly colored comic book panel, we see two characters engaging in a lively conversation. One character, with a \textcolor{critTextB}{bright yellow head} and \textcolor{critTextC}{a red shirt}, exclaims, ``An inventor? I don't believe you!'' His name, Marco, is displayed in a white rectangle. Opposite him, another character, sporting a black hairstyle and \textcolor{critTextC}{a blue shirt}, confidently asks, ``Why not?'' His name, Babi, \textcolor{critTextA}{is also shown in a white rectangle}. The contrasting colors of their clothes and hair add to the dynamic energy of their exchange, as they stand against a plain blue backdrop. The speech bubbles capture their dialogue, highlighting the playful and inquisitive spirit of their interaction.
\\

\arrayrulecolor{critB}
\midrule
\arrayrulecolor{grid}
Ground Truth &
\vspace{-0.95em}
\begin{tabular}{p{8cm} p{3cm}}
\cellcolor{critA} Completeness and Coverage & A $>$ B \\
\cellcolor{critB} Visual Grounding and Details & B $>$ A \\
\cellcolor{critC} Factuality / No Hallucination & A $>$ B \\
\end{tabular}
\\

\bottomrule
\end{tabular}
}
\captionof{table}{Example of an open-ended evaluation instance for blind storytelling. Response A conveys the \textcolor{critTextA}{moral of the story} and fully \textcolor{critTextA}{follows the instruction to avoid mentioning rectangles}. Response B includes more \textcolor{critTextB}{visual details} but \textcolor{critTextC}{hallucinates the shirt colors}.}
\label{tab:example_openended_conflict_blind}
\end{minipage}
\end{table*}

\begin{table*}[!ht]
\begin{minipage}{1.0\textwidth}
\centering
\scalebox{0.98}{
\renewcommand{\arraystretch}{1.15}
\setlength{\tabcolsep}{6pt}
\begin{tabular}{l p{11cm}}
\rowcolor{tbhead}
\toprule
\multicolumn{2}{l}{\bfseries Open-ended Example 2} \\
\midrule
& \includegraphics[width=8.2cm]{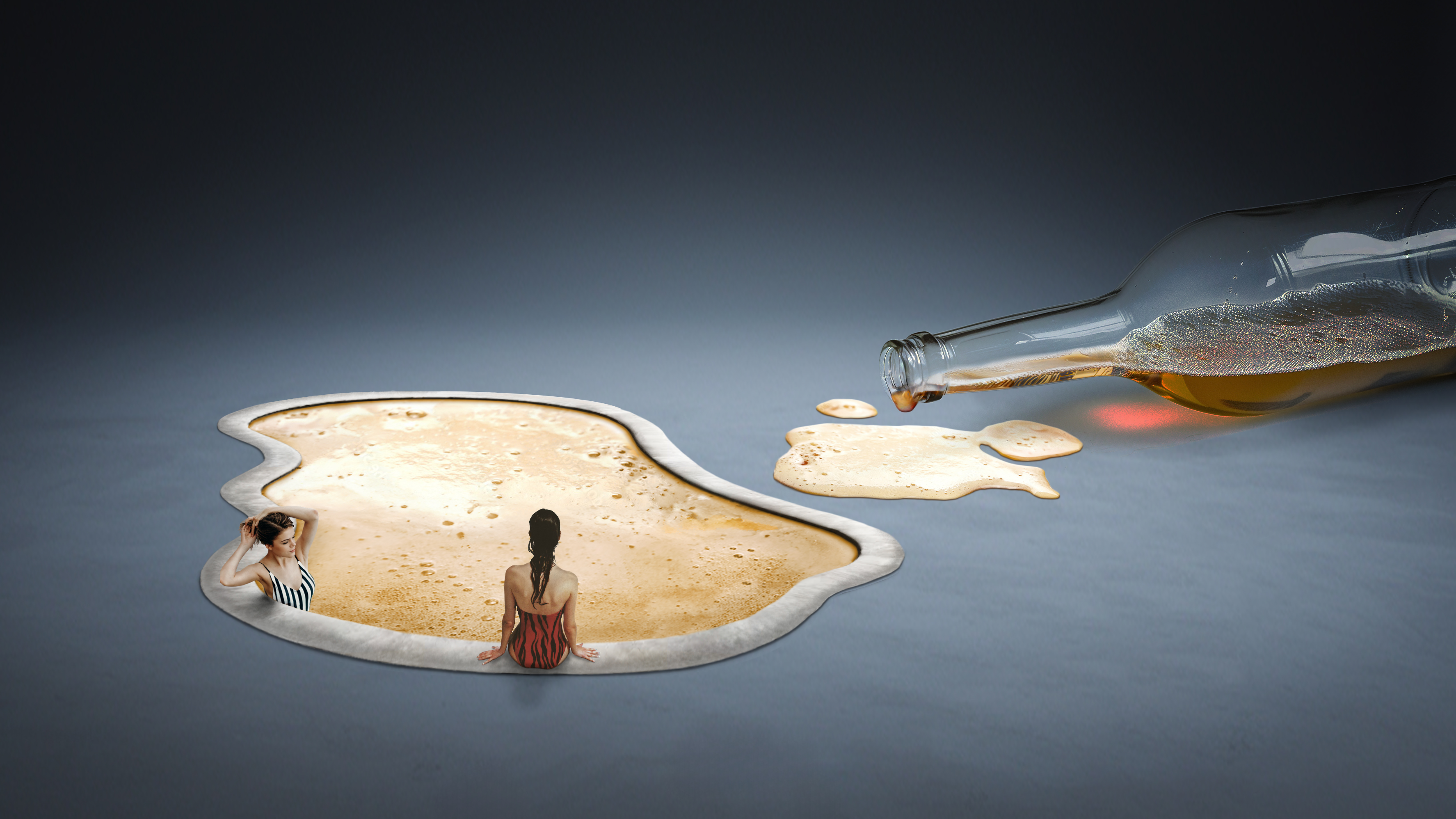} \\[0.3em]
Question &
What does this image show?
\\
\midrule
\arrayrulecolor{grid}

\rowcolor{tbshade}
\makecell[lt]{Response A \\ (Llama-3.2-11B-Vision-Instruct)} &
The image depicts a surreal scene where a bottle of beer is pouring its contents onto the ground, creating a puddle that morphs into a miniature pool. Within this pool, two women are shown sitting and posing in swimsuits, as if they were enjoying a day at the beach or poolside. \textcolor{critTextE}{The juxtaposition of the beer’s spillage with the serene atmosphere of the women creates a humorous and unexpected visual contrast.}
\\
\midrule
\makecell[lt]{Response B \\ (Qwen2.5-VL-7B-Instruct)} &
\textcolor{critTextD}{This image is a creative and surreal depiction that combines elements of everyday objects with imaginative scenarios.} It shows a large, tilted bottle pouring \textcolor{critTextD}{liquid} onto the ground. The \textcolor{critTextD}{liquid} spreads out in a pool around \textcolor{critTextD}{two women} who are sitting on the edge of the puddle. \textcolor{critTextD}{One woman} is partially submerged in the liquid, while \textcolor{critTextD}{the other} \textcolor{critTextC}{sits on the edge looking at her}. The scene is contrasted by the dark, plain \textcolor{critTextD}{background}, which makes the bottle and the women stand out prominently. This type of imagery is often used in \textcolor{critTextA}{advertising or artistic representations} to convey a message or evoke emotions.
\\
\arrayrulecolor{critD}\midrule\arrayrulecolor{grid}
Ground Truth &
\vspace{-0.95em}
\begin{tabular}{p{7cm} p{3cm}}
\cellcolor{critA} Completeness and Coverage & B $>$ A \\
\cellcolor{critC} Factuality / No Hallucination & A  $>$ B \\
\cellcolor{critE} Creativity and Expressiveness & A $>$ B\\
\cellcolor{critD} Clarity and Coherence & B $>$ A \\
\end{tabular}
\\

\bottomrule
\end{tabular}
}
\captionof{table}{Example of an open-ended evaluation instance for creative-image captioning. Response A creatively highlights the \textcolor{critTextE}{humorous and unexpected effect} and introduces \textcolor{critTextC}{less hallucinated content}. Response B includes a brief \textcolor{critTextD}{TL;DR} and an \textcolor{critTextA}{inference} about image usage, with a more \textcolor{critTextD}{coherent flow across elements}. }
\label{tab:example_openended_conflict_beerpool}
\end{minipage}
\end{table*}



\begin{table*}[!ht]
\begin{minipage}{1.0\textwidth}
\centering
\scalebox{0.99}{
\renewcommand{\arraystretch}{1.15}
\setlength{\tabcolsep}{3pt}
\begin{tabular}{p{2.1cm} p{14cm}}
\rowcolor{tbhead}
\toprule
\multicolumn{2}{l}{\bfseries Open-ended Example 3} \\
\midrule
& \includegraphics[width=4.8cm]{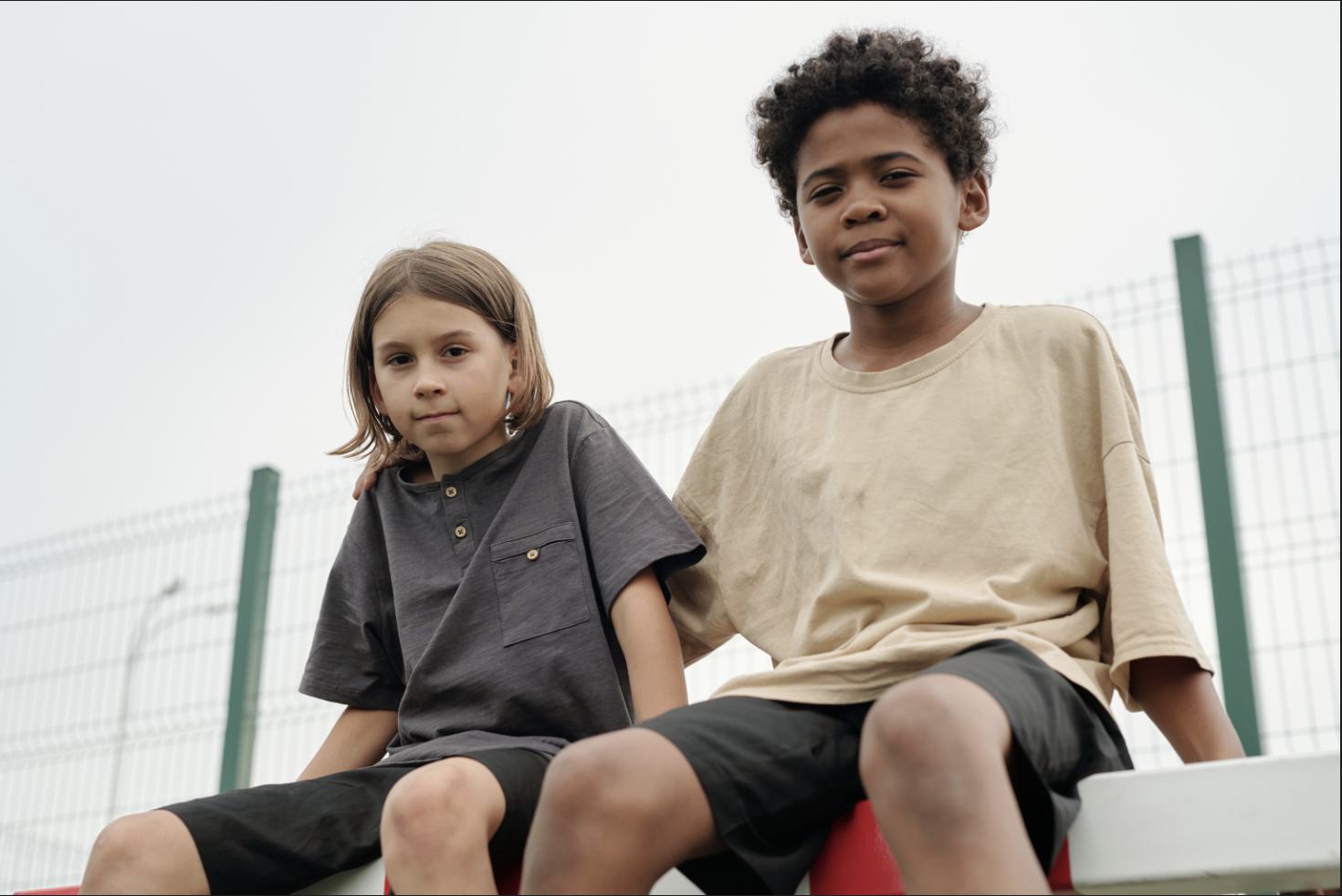} \\[0.3em]
Question &
Create detailed one sentence image caption, capture the essence and include the following structured elements without assumptions or ambiguous language: Background and Foreground: Separate background objects and foreground objects to enhance focus;
Location: The setting can be either indoors or outdoors; Shot Size: The image sizes: Extreme close up, Close up, Medium close up, Medium shot, Medium full shot, Full length. Use this parameter if a person is included in the photo; Angle: low angle shot, high angle shot, top view shot, flat lay. If the angle is straight, omit this specification; Temperature: Color temperature should be considered, with options including cold, warm, neutral, or vibrant. If the temperature is neutral, you may choose to not specify this detail; Lens: Use a fisheye lens or wide-angle as appropriate. If neither is used, this detail can be omitted; Lighting: Very important options include High-key Lighting, Low-key Lighting, Dramatic light, Backlight, silhouette, cinematic light; Composition: Employ techniques such as the rule of thirds, double exposure, over-the-shoulder shots, or symmetrical composition to structure the image; Day Time: Specify the time of day (e.g., blue hour, sunrise, morning, noon, afternoon) if this parameter is explicitly relevant; \textcolor{critTextA}{Person Caption:} If a person is included, describe their race of skin color (Asian, Caucasian, Black, Arab, Indian, etc), hair (black, brown, blond, white/gray, and rarely red) view direction, emotions and mood, and clothing details.
\\
\midrule
\arrayrulecolor{grid}

\rowcolor{tbshade}
\makecell[lt]{Response A \\ (GPT-4.1)} &
Foreground: Two children, one with light skin and straight brown hair, wearing a charcoal grey shirt and black shorts, and the other with dark skin and short curly black hair, dressed in a beige shirt and black shorts, are seated close to each other with the child on the right resting an arm around the other’s shoulders; both appear calm and friendly, looking directly at the camera. Background: \textcolor{critTextB}{Metal fencing with vertical green posts is visible, along with an overcast sky and faint outlines of a streetlight or utility pole.} Location: Outdoors. Shot Size: Medium shot. Angle: Low angle shot. Temperature: Neutral. Lighting: Natural diffused daylight. Composition: Centered composition, subjects positioned side by side with upper bodies and faces as focal point. Day Time: Daytime.
\\
\midrule
\makecell[lt]{Response B \\ (GPT-4.1)} &
\textcolor{critTextD}{Foreground: Two boys are sitting close together on what appears to be outdoor seating;} Background: A wire mesh fence and holding poles are visible against a blank, overcast sky; Location: Outdoors; Shot Size: Medium shot; Angle: Low angle shot; Temperature: Neutral; Lighting: Diffused natural light due to an overcast day; \textcolor{critTextD}{Composition: Symmetrical composition with both boys centered in the frame}; Day Time: Daytime, likely noon or afternoon given the soft daylight; \textcolor{critTextA}{Person Caption:} On the left, a \textcolor{critTextA}{Caucasian} boy with straight, shoulder-length brown hair faces the camera with a neutral, slightly guarded expression, wearing a charcoal grey henley shirt and black shorts; on the right, a Black boy with short, curly black hair also faces the camera with a calm, faintly smiling demeanor, dressed in a loose beige t-shirt and black shorts, his \textcolor{critTextC}{left arm draped supportively behind the other boy.}
\\
\arrayrulecolor{critD}\midrule\arrayrulecolor{grid}

Ground Truth &
\vspace{-0.95em}
\begin{tabular}{p{7cm} p{3cm}}
\cellcolor{critA} Completeness and Coverage & B $>$ A \\
\cellcolor{critB} Visual Grounding and Details & A $>$ B \\
\cellcolor{critC} Factuality / No Hallucination & A $>$ B \\
\cellcolor{critD} Clarity and Coherence & B $>$ A\\
\end{tabular}
\\

\bottomrule
\end{tabular}
}
\vspace{-1mm}
\captionof{table}{Example of an evaluation instance in instruction-rich image analysis. Both responses are randomly sampled from GPT-4.1. Response A provides more \textcolor{critTextB}{background details} and is \textcolor{critTextC}{less hallucinated}, while Response B better follows \textcolor{critTextA}{all required elements in the user prompt} and offers a more \textcolor{critTextD}{clear and readable description of the foreground and composition}.}
\label{tab:example_openended_conflict_kids}
\end{minipage}
\end{table*}

%

\begin{table*}[!ht]
\begin{minipage}{1.0\textwidth}
\centering
\scalebox{0.97}{
\renewcommand{\arraystretch}{1.1}
\setlength{\tabcolsep}{5pt}
\begin{tabular}{p{2.4cm} p{14cm}}
\rowcolor{tbhead}
\toprule
\multicolumn{2}{l}{\bfseries Verifiable Reasoning Example 1} \\
\midrule
& \includegraphics[width=4cm]{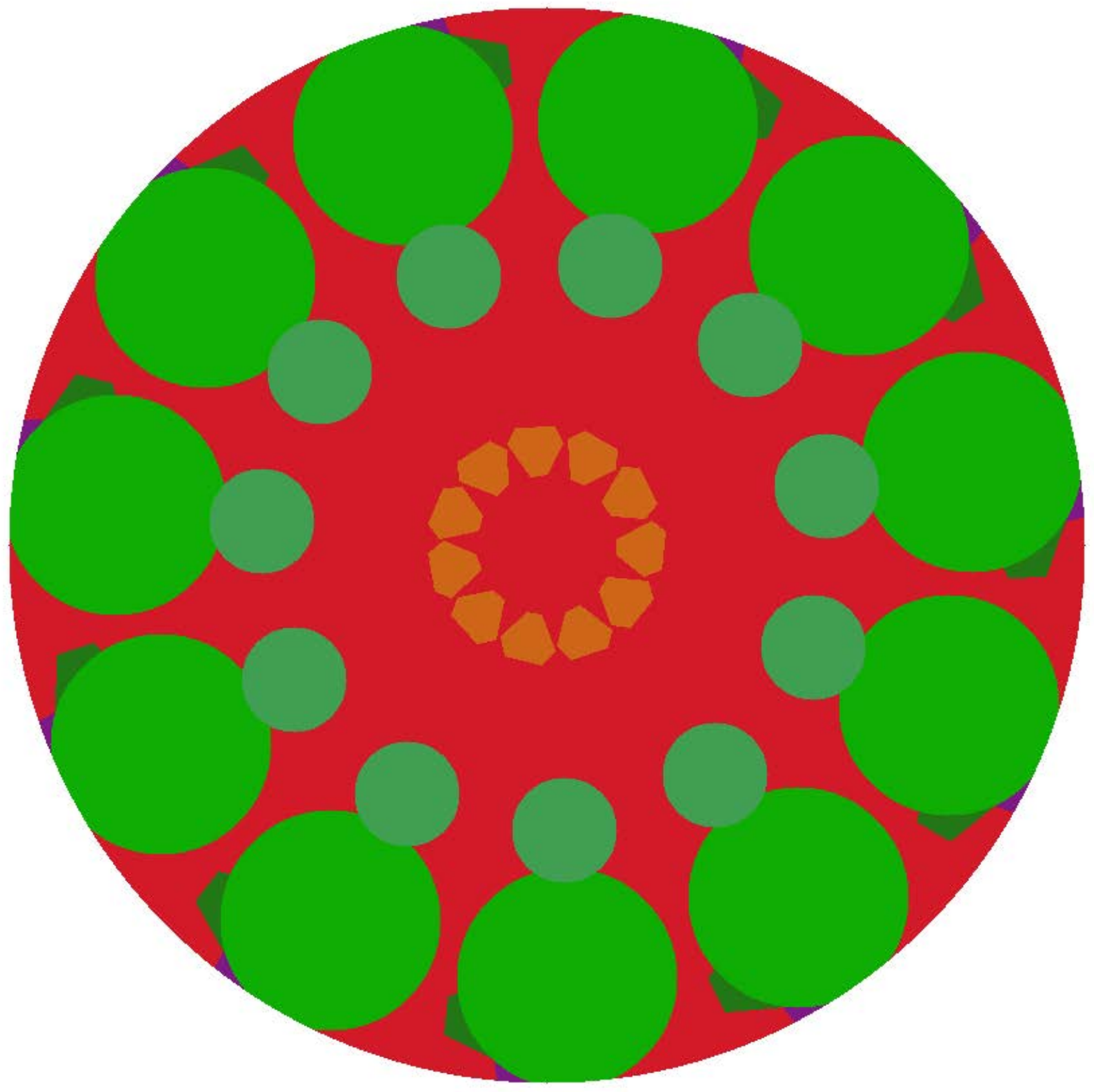} \\[0.3em]
Question &
The entire pie seen in the figure is divided among several children. Each child receives a piece of pie, and each piece of pie looks identical. The maximum possible number of children there is: Choices: (A) 10 (B) 12 (C) 9 (D) 11. \; 
\\
Answer & D \\
\arrayrulecolor{critC}\midrule\arrayrulecolor{grid}

\rowcolor{tbshade}
\makecell[lt]{Response A \\ (Qwen2.5-VL\\-72B-Instruct)} &
To determine the maximum possible number of children who can receive an identical piece of the pie, we need to analyze the symmetry of the figure. 1) Identify the Symmetry: The figure appears to be a circular design with a central red area and green circles arranged around it. The key is to find the rotational symmetry of the design. 2) Count the green circles: There are 12 green circles arranged around the central red area. This suggests that the design might have 12-fold rotational symmetry. 3) If we rotate the figure by \( \frac{360^\circ}{12} = 30^\circ \), the figure should look the same. This means that the design is symmetric under a 12-fold rotation. 4) Conclusion: Since the design has 12-fold rotational symmetry, the pie can be divided into 12 identical pieces. Therefore, the maximum possible number of children who can each receive an identical piece of the pie is 12. Thus, the correct answer is \(\boxed{B}\).
\\
\midrule
\makecell[lt]{Response B \\ (Gemini-2.5\\-Flash)} &
To find the maximum possible number of children, we need to determine the number of identical pieces the pie can be divided into. This means identifying the smallest repeating unit in the circular pattern.
Let's observe the elements arranged in a circle:
1) Large green circles: Count the number of large green circles arranged in the outer ring. There are 12 of them.
2) Small green/grey circles: Count the number of smaller green/grey circles in the inner ring. There are also 12 of these.
\textcolor{critTextB}{3) Orange shapes: Count the number of orange shapes in the very center.} There are 12 of these as well.

Since all three types of repeating elements (large green circles, small green/\textcolor{critTextC}{grey} circles, and orange shapes) appear 12 times and are arranged symmetrically around the center, the pie exhibits 12-fold rotational symmetry.
If each child receives an identical piece of pie, then the pie must be divided into sectors that are identical in shape and contain identical patterns. The smallest such sector would be one that contains one of each of the repeating elements.
\textcolor{critTextD}{Imagine drawing} radial lines from the center of the pie that equally divide the circle. If we divide the pie into 12 equal sectors (each 360/12 = 30 degrees), each sector will contain: One large green circle. One small green/grey circle. One orange shape.

Since these 12 pieces would be identical and cover the entire pie, the maximum possible number of children is 12. \textcolor{purple}{If there were fewer children (e.g., 6), each piece would be larger and contain multiple sets of these elements (e.g., two large green circles, two small green circles, two orange shapes), but the question asks for the *maximum* number, which corresponds to the smallest possible identical pieces.}
Therefore, the maximum number of children is 12.
The final answer is $\boxed{\text{12}}$.
\\
\arrayrulecolor{critD}\midrule\arrayrulecolor{grid}

Ground Truth &
\vspace{-0.95em}
\begin{tabular}{p{7cm} p{3cm}}
\cellcolor{critB} Visual Grounding & B $>$ A \\
\cellcolor{critD} Logical Coherence and Consistency & A $>$ B \\
\cellcolor{critC} Factual Correctness / No Hallucination & A $>$ B \\
\cellcolor{critE} Reflection and Exploration & B $>$ A \\
\cellcolor{critF} Conciseness and Efficiency & A $>$ B \\
\end{tabular}
\\

\bottomrule
\end{tabular}
}
\vspace{-1mm}
\captionof{table}{Example of an evaluation instance for math reasoning.
Both responses appeal to rotational symmetry but arrive at incorrect final answers. Response A is more \textcolor{critTextF}{concise} and \textcolor{critTextC}{avoid hallucinated colors}. Response B demonstrates slightly better \textcolor{critTextB}{grounding for the inner orange shapes} and actively explores \textcolor{critTextE}{alternative solutions}, yet contains some \textcolor{critTextD}{logical shifts} that reduce coherence.}
\label{tab:verifiable_reasoning_symmetry}
\end{minipage}
\end{table*}

\begin{table*}[!ht]
\begin{minipage}{1.0\textwidth}
\centering
\renewcommand{\arraystretch}{1.15}
\setlength{\tabcolsep}{6pt}
\begin{tabular}{l p{12cm}}
\rowcolor{tbhead}
\toprule
\multicolumn{2}{l}{\bfseries Verifiable Reasoning Example 2} \\
\midrule
& \includegraphics[width=6.2cm]{figures/appendix/conflict/reasoning-104.png} \\[0.3em]
Question &
Is the number of green bus greater than the number of blue school bus? Options: (A) Yes (B) No. \; 
\\
Answer &  A \\
\arrayrulecolor{critC}\midrule\arrayrulecolor{grid}

\rowcolor{tbshade}
\makecell[lt]{Response A \\ (Gemini-2.5-Flash)} &
1.Identify green buses: In the image, there is one long green vehicle that is a bus. So, the number of green buses is 1.
2.Identify blue school buses: There are no blue vehicles visible in the image, and certainly no blue school buses. So, the number of blue school buses is 0.
3. Compare the numbers: 1 (green buses) is greater than 0 (blue school buses).
Therefore, the number of green buses is greater than the number of blue school buses.
The final answer is $\boxed{A}$
\\
\midrule
\makecell[lt]{Response B \\ (Gemini-2.5-Flash)} &
Here's a breakdown of the vehicles in the image:
\textcolor{critTextE}{-} There is one green bus in the \textcolor{critTextB}{lower right} portion of the image.
\textcolor{critTextE}{-} There is a \textcolor{critTextB}{gray dirt bike in the middle-left}.
\textcolor{critTextE}{-} There is a \textcolor{critTextB}{golden motorcycle} (possibly a chopper or custom bike) with red and green accents in the upper-middle.
\textcolor{critTextE}{-} There is a gray airplane with blue and green accents in the upper-right.
Based on this observation:
- Number of green buses = 1
- Number of blue school buses = 0 (\textcolor{critTextE}{There are no blue buses}, \textcolor{critTextC}{nor any vehicles that distinctly look like a school bus of any color}, let alone blue.)
Since 1 is greater than 0, the number of green buses is greater than the number of blue school buses.
The final answer is $\boxed{A}$
\\
\arrayrulecolor{critB}\midrule\arrayrulecolor{grid}

Ground Truth &
\vspace{-0.95em}
\begin{tabular}{p{7cm} p{3cm}}
\cellcolor{critB} Visual Grounding & B $>$ A \\
\cellcolor{critC} Factual Correctness / No Hallucination & A $>$ B\\
\cellcolor{critE} Reflection and Exploration & B $>$ A \\
\cellcolor{critF} Conciseness and Efficiency & A $>$ B \\
\end{tabular}
\\

\bottomrule
\end{tabular}
\captionof{table}{Example of an evaluation instance for object counting. Both Gemini-2.5-Flash responses correctly identify more green buses than blue ones. Response A is more \textcolor{critTextF}{compact}. Response B \textcolor{critTextB}{provides finer object details and spatial relationships} and \textcolor{critTextE}{explicitly checks each visible object}, but introduces a minor \textcolor{critTextC}{hallucination by stating that no school bus is present}, even though one appears in the image.}
\label{tab:verifiable_reasoning_buses}
\end{minipage}
\end{table*}

\begin{table*}[!ht]
\begin{minipage}{1.0\textwidth}
\centering
\makebox[\textwidth][c]{
\scalebox{0.975}{
\renewcommand{\arraystretch}{1.03}
\setlength{\tabcolsep}{2pt}
\begin{tabular}{p{2cm} p{14.9cm}}
\rowcolor{tbhead}
\toprule
\multicolumn{2}{l}{\bfseries Verifiable Reasoning Example 3} \\
\midrule
& \includegraphics[width=6.75cm]{figures/appendix/conflict/reasoning-57.png} \\
[-1.3em]

Question &
In the diagram of the food web shown, a pointing arrow from A to B indicates that A is the source of energy that helps B survive. If the number of lizard and grasshopper decrease, the population of grass will likely?  (A) decreases (B) increases (C) remains the same (D) cannot be determined from the image.\;
\\
Answer & B \\
\arrayrulecolor{critC}\midrule\arrayrulecolor{grid}

\rowcolor{tbshade}
\makecell[lt]{Response A \\ (MiMo-VL\\-7B-RL)} &
\small
\texttt{<think>} I need to analyze this food web diagram to understand the relationships between organisms and determine what would happen to the grass population if lizards and grasshoppers decrease.
\textcolor{critTextB}{
Looking at the diagram:
- Grass is eaten by: grasshopper, rabbit, and mouse
- Grasshopper is eaten by: lizard and hawk
- Lizard is eaten by: hawk
- Rabbit is eaten by: hawk
- Mouse is eaten by: hawk and snake
- Snake is eaten by: hawk
}
If the number of lizards and grasshoppers decreases:
1. Grasshoppers eat grass, so fewer grasshoppers would mean less consumption of grass.
2. Lizards don't directly eat grass, but they eat grasshoppers. With fewer lizards, there would be fewer predators controlling the grasshopper population, which could potentially lead to more grasshoppers in the long run. However, the question states that both lizard and grasshopper populations are decreasing.
Since both grasshoppers (which eat grass) and lizards (which control grasshoppers) are decreasing, we have two opposing effects:
- Fewer grasshoppers means less grass consumed
- Fewer lizards means less predation on grasshoppers, which could lead to more grasshoppers in the future.
However, since the question specifically asks about the immediate effect of these decreases, the most direct impact would be that with fewer grasshoppers consuming grass, the grass population would likely increase. \texttt{</think>} 
 In this ecosystem:
 - Grass is consumed by three primary herbivores: grasshoppers, rabbits, and mice
 - Grasshoppers are preyed upon by both lizards and hawks
 - Lizards are preyed upon by hawks. 
 When the number of lizards and grasshoppers decreases:
 1. Grasshoppers directly consume grass. With fewer grasshoppers, there would be less consumption of grass.
 2. Lizards don't directly consume grass, but they do control the grasshopper population through predation. With fewer lizards, there would be less predation pressure on grasshoppers, which could potentially lead to more grasshoppers in the future. However, the question specifies that both lizard and grasshopper populations are decreasing. The most immediate effect would be reduced grass consumption due to fewer grasshoppers. Since grass is also consumed by rabbits and mice (whose populations aren't mentioned as changing), the reduction in grasshoppers would likely result in increased grass availability. Therefore, the population of grass will likely \boxed{(B) increases}.
\\
\midrule
\makecell[lt]{Response B \\ (Gemini-2.5\\-Flash)} &
\small
To determine the likely change in the grass population, we need to look at what organisms consume grass in this food web.
1. Identify consumers of Grass:
The diagram shows arrows from Grass to Grasshopper, Rabbit, and Mouse. This means Grasshoppers, Rabbits, and Mice all eat Grass.
2. Analyze the given scenario:
“the number of lizard and grasshopper decrease”.
3. Focus on the direct impact on Grass:
-- The Grasshopper population decreases. Since grasshoppers eat grass, a decrease in their numbers means there will be fewer organisms consuming grass.
-- The Lizard population decreases. Lizards eat grasshoppers. \textcolor{critTextE}{While a decrease in lizards might lead to an increase in grasshoppers (their prey) in some scenarios, the problem explicitly states that grasshoppers \emph{also} decrease. Therefore, the decrease in lizards doesn't counteract the effect of fewer grasshoppers eating grass.} In fact, fewer lizards might be a consequence of fewer grasshoppers.
4. Conclusion:
With fewer grasshoppers eating grass, the predation pressure on the grass population is reduced. Assuming other factors (like Rabbit and Mouse populations, or environmental conditions) remain constant, the grass will be consumed less. This reduced consumption will likely lead to an increase in the grass population.
Therefore, if the number of lizard and grasshopper decrease, the population of grass will likely increase. The final answer is \boxed{B}.\\
\arrayrulecolor{critD}\midrule\arrayrulecolor{grid}
Ground Truth &
\vspace{-0.95em}
\begin{tabular}{p{7cm} p{3cm}}
\cellcolor{critB} Visual Grounding & A $>$ B \\
\cellcolor{critE} Reflection and Exploration & B $>$ A \\
\cellcolor{critF} Conciseness and Efficiency & B $>$ A\\
\end{tabular}
\\

\bottomrule
\end{tabular}
}
}

\vspace{-1.5mm}
\captionof{table}{Example of an evaluation instance in judging biological reasoning. Both responses reach the correct conclusion (grass increases). Response A provides a more complete and detailed \textcolor{critTextB}{description of the food web}. Response B is more \textcolor{critTextF}{efficient} and together shows \textcolor{critTextE}{deeper reflection} by explicitly recognizing that the decrease in lizards \textcolor{critTextE}{does not counteract the effect} of fewer grasshoppers eating grass.}
\label{tab:verifiable_reasoning_foodweb}
\end{minipage}
\end{table*}

%



\clearpage
\clearpage

\section{Additional Experimental Results}
\label{sec:appendix-experients}

\subsection{To Think or Not?}

To study how thinking influences judge model behaviors in following diverse criteria, we examine two strong open-source LMM families that are known for producing high-quality reasoning traces across diverse domains.
InternVL3.5~\cite{wang2025internvl3.5} supports both thinking (default) and non-thinking modes controlled via system prompts, enabling a clean mode-switch comparison within the same model weights. Qwen3-VL~\cite{yang2025qwen3} provides two model variants—Instruct and Thinking—the latter specifically developed to enhance reasoning capabilities.

Table~\ref{tab:append-results-thinking} presents the results, showing two trends:

\begin{enumerate}[itemsep=5pt]
    \item \textit{Smaller models benefit more from thinking, while larger models show limited gains.} \\[2pt]
    For the smaller 8B models, both the mode-switch (InternVL3.5) and model-variant (Qwen3-VL) show clear and consistent improvements from enabling thinking, with gains across all metrics on both open-ended and reasoning splits. 
    In contrast, the larger $\sim$30B models show more modest effects: thinking brings no significant improvement for InternVL3.5, and for Qwen3-VL it enhances reasoning judgments but slightly reduces performance on open-ended splits. 
    
    These results suggest that explicit thinking during judgment helps smaller models better adhere to fine-grained evaluation criteria and produce criterion-specific critic reasoning. For larger models that already internalize such criteria-following capacities and reasoning patterns, additional thinking offers little benefit and may even amplify judge model biases.

    \item \textit{Thinking benefits reasoning judgments more.} \\[2pt]
    For both Qwen3-VL and InternVL families, the thinking judge yields larger gains in evaluating verifiable reasoning than in open-ended generation tasks. Specifically, \textit{Qwen3-VL-Thinking} outperforms its \textit{Instruct} variant at both 8B and 32B scales, demonstrating improved ability to recognize criterion trade-offs and capture preference conflicts.
    This finding differs from the observation in Sec.~\ref{subsec:experiment-additional-discussion}, where RL-finetuned reasoning models on domain-specific data based on Qwen2.5-VL showed no improvement in reasoning judgments and even weakened trade-off recognition.
    
    This indicates that the capacity for reasoning judgment arises from general thinking abilities developed through broad reasoning training across diverse domains, rather than from narrow domain-specific fine-tuning, which often leads to overfitting and reduced generalization.
\end{enumerate}

\begin{table*}[!htp]
\centering
\setlength{\tabcolsep}{3pt}
\renewcommand{\arraystretch}{1.15}

\begin{tabular}{l|cccc|cccc}
\toprule
\multirow{2}{*}{Model} &
\multicolumn{4}{c|}{Open-ended} &
\multicolumn{4}{c}{Verifiable Reasoning} \\
\cmidrule(lr){2-5} \cmidrule(lr){6-9}
& Pluralistic & Conflict & Tradeoff & Crit.-Avg. 
& Pluralistic & Conflict & Tradeoff & Crit.-Avg. \\
\midrule

InternVL3.5-8B (no-think) 
& 23.41 & 24.95 & 44.66 & 59.09
& 27.78 & 28.83 & 54.13 & 61.96
\\

InternVL3.5-8B (think)
& 25.08 & 32.34 & 62.14 & 61.04
& 32.54 & 39.15 & 69.72 & 65.85
\\
\rowcolor{gray!6}
\small$\Delta$
& \small{\textcolor{ForestGreen}{+1.67}} 
& \small{\textcolor{ForestGreen}{+7.39}} 
& \small{\textcolor{ForestGreen}{+17.48}} 
& \small{\textcolor{ForestGreen}{+1.95}}
& \small{\textcolor{ForestGreen}{+4.76}} 
& \small{\textcolor{ForestGreen}{+10.32}} 
& \small{\textcolor{ForestGreen}{+15.59}} 
& \small{\textcolor{ForestGreen}{+3.89}}
\\

Qwen3-VL-8B-Instruct
& 18.39 & 18.36 & 34.47 & 54.16
& 16.67 & 19.22 & 35.78 & 56.92
\\

Qwen3-VL-8B-Thinking
& 24.75 & 37.33 & 66.50 & 60.61
& 38.89 & 51.25 & 83.49 & 70.70
\\
\rowcolor{gray!6}
\small$\Delta$
& \small{\textcolor{ForestGreen}{+6.36}}
& \small{\textcolor{ForestGreen}{+18.97}}
& \small{\textcolor{ForestGreen}{+32.03}}
& \small{\textcolor{ForestGreen}{+6.45}}
& \small{\textcolor{ForestGreen}{+22.22}}
& \small{\textcolor{ForestGreen}{+32.03}}
& \small{\textcolor{ForestGreen}{+47.71}}
& \small{\textcolor{ForestGreen}{+13.78}}
\\

\midrule

InternVL3.5-38B (no-think)
& 29.43 & 35.93 & 61.65 & 65.36
& 40.48 & 47.69 & 75.23 & 71.37
\\

InternVL3.5-38B (think)
& 30.43 & 33.73 & 64.08 & 65.10
& 37.30 & 47.69 & 75.23 & 69.82
\\
\rowcolor{gray!6}
\small$\Delta$
& \small{\textcolor{ForestGreen}{+1.00}}
& \small{\textcolor{BrickRed}{-2.20}}
& \small{\textcolor{ForestGreen}{+2.43}}
& \small{\textcolor{BrickRed}{-0.26}}
& \small{\textcolor{BrickRed}{-3.18}}
& \small{\textcolor{gray}{0.00}}
& \small{\textcolor{gray}{0.00}}
& \small{\textcolor{BrickRed}{-1.55}}
\\

Qwen3-VL-32B-Instruct
& 30.43 & 40.32 & 68.93 & 65.49

& 39.68 & 48.75 & 70.64 & 71.42
\\

Qwen3-VL-32B-Thinking
& 29.10 & 40.12 & 67.96 & 64.88
& 43.65 & 53.38 & 80.73 & 73.88
\\
\rowcolor{gray!6}
\small$\Delta$
& \small{\textcolor{BrickRed}{-1.33}}
& \small{\textcolor{BrickRed}{-0.20}}
& \small{\textcolor{BrickRed}{-0.97}}
& \small{\textcolor{BrickRed}{-0.61}}
& \small{\textcolor{ForestGreen}{+3.97}}
& \small{\textcolor{ForestGreen}{+4.63}}
& \small{\textcolor{ForestGreen}{+10.09}}
& \small{\textcolor{ForestGreen}{+2.46}}
\\

\bottomrule
\end{tabular}

\caption{Comparison of thinking vs. non-thinking LMM judges. Relative improvements are shown in \textcolor{ForestGreen}{green} and decreases in \textcolor{BrickRed}{red}.}
\label{tab:append-results-thinking}
\end{table*}

%


\subsection{Joint Multi-Criterion Judgment}
\label{subsec:appendix-joint-eval}

\paragraph{Task Formulation.} In our standard setting, each evaluation instance (prompt + response pair) with $K$ applicable criteria requires the LMM judge to perform $K$ separate inferences, each instantiated with a single-criterion evaluation prompt as shown in Table~\ref{tab:append_prompt_template}. Here, we investigate an alternative setting in which the judge performs \textit{joint multi-criterion judgment}. For each instance, all $K$ criteria are presented to the model simultaneously, and a \emph{single} inference is used to produce criterion-level judgments for all criteria. Table~\ref{tab:joint_prompt_template} shows the corresponding joint evaluation prompt: the model is instructed to treat each criterion separately while outputting all criterion-level judgments in one pass. All inference hyper-parameters remain identical to the standard single-criterion setup, allowing us to isolate the effect of joint prompting on judge model behavior.

\paragraph{Results and Analysis.} We evaluate four top-performing proprietary LMMs—GPT-4o, GPT-5, o4-mini, and Gemini-2.5-Pro—under the joint multi-criterion judgment setting. 
The results in Table~\ref{tab:appendix-experiment-joint-criterion} reveal two major observations:

\begin{enumerate}[itemsep=5pt,topsep=3pt]
    \item \textit{Joint multi-criterion judgment affects models unevenly.}  \\
    GPT-4o exhibits a clear performance drop across all metrics, suggesting that combining multiple criteria in a single pass amplifies its internal biases and weakens its ability to follow diverse criteria. GPT-5, in contrast, benefits from the joint setting—showing improved pluralistic accuracy and better alignment with human preferences on conflict cases.  For o4-mini and Gemini-2.5-Pro, the effects of joint criterion judgment are mixed on open-ended tasks but consistently drop on reasoning domains.

    \item \textit{Joint judgment generally reduces the model’s sensitivity to criterion-level trade-offs.} \\
    Across all models—except GPT-5 in the reasoning split—the trade-off sensitivity decreases.
    This trend is expected: given the autoregressive nature of LLMs/LMMs, generating multiple criterion-level judgments within a single inference pass inevitably induces inter-criterion dependencies, making the judge model more likely to assign the same preference direction across criteria for the same pair of responses. Table~\ref{tab:openended_age_hair_conflict_batch_example} shows a case where \textit{o4-mini} fails to capture criterion-level conflicts under joint criterion judgment.
\end{enumerate}

\clearpage

\begin{table*}[!ht]
\centering
\begin{minipage}{1.0\textwidth}
\vspace{0mm}
\centering
\begin{tcolorbox}
\centering
\begin{tabular}{p{0.96\textwidth}}
You are an expert in evaluating the quality of AI-generated responses according to multiple evaluation criteria. Your task is to assess two responses generated by different AI assistants in reply to a user's question about an image. The image is provided as part of the input. \\[0.6ex]
You must evaluate the responses based on the following \texttt{\{K\}} evaluation criteria. \textbf{Analyze each criterion independently and exclusively}---your judgment on one criterion should not influence your judgment on another. Do not consider any other dimensions or criteria beyond what is specified below. \\[0.6ex]
\texttt{\{Criteria\}} \\[0.6ex]
Here are the inputs for your evaluation: \\
\mbox{}\quad [Question]: \texttt{\{Question\}} \\
\mbox{}\quad [Response 1]: \texttt{\{Response1\}} \\
\mbox{}\quad [Response 2]: \texttt{\{Response2\}} \\[0.6ex]
\textbf{Instructions:} For \emph{each} criterion listed above, you must: (1) provide a detailed justification for your evaluation, referring to specific elements in the responses, how they align with that criterion, and relevant visual details from the image; and (2) on the final line of each criterion's evaluation, provide your judgment based solely on that criterion. You must choose one response as better; do not indicate a tie. \\[0.6ex]
Format your response as follows for each criterion: \\[0.6ex]
\mbox{}\quad \textbf{Criterion: [Criterion Name 1]} \\
\mbox{}\quad [Your detailed justification here] \\
\mbox{}\quad Judgment: Response X is better. \\[0.3ex]
\mbox{}\quad \textbf{Criterion: [Criterion Name 2]} \\
\mbox{}\quad [Your detailed justification here] \\
\mbox{}\quad Judgment: Response X is better. \\[0.3ex]
\mbox{}\quad \dots (continue for all criteria) \\[0.6ex]
Strictly follow this format: \texttt{``Response X is better.''} on the last line of each criterion block.
\end{tabular}
\end{tcolorbox}
\vspace{-2.5mm}
\caption{Joint multi-criterion evaluation prompt for LMM judges. All applicable criteria for an evaluation instance are assessed in a single pass, and the judge is explicitly instructed to treat each criterion separately.}
\label{tab:joint_prompt_template}
\end{minipage}
\end{table*}

\begin{table*}[t]
\centering
\renewcommand{\arraystretch}{1.1}
\begin{tabular}{l|cccc|cccc}
\toprule
\multirow{2}{*}{\centering Model} & \multicolumn{4}{c|}
{Open-ended} & \multicolumn{4}{c}{Verifiable Reasoning} \\
\cmidrule(lr){2-5}\cmidrule(lr){6-9}
 &

Pluralistic & Conflict & Tradeoff & Criterion Avg. &
Pluralistic &  Conflict & Tradeoff & Criterion Avg. \\
\midrule

GPT-4o
& 31.44 & 44.91 & 66.02 & 69.57
& 41.27 & 55.16 & 84.40 & 69.79
\\

+ Joint multi-criterion
& 30.10 & 26.35 & 38.83 & 64.91
& 34.13 & 34.52 & 58.72 & 63.15
\\

\rowcolor{gray!6}
\small$\Delta$
& \small{\textcolor{BrickRed}{-1.34}}
& \small{\textcolor{BrickRed}{-18.56}}
& \small{\textcolor{BrickRed}{-27.19}}
& \small{\textcolor{BrickRed}{-4.66}}
& \small{\textcolor{BrickRed}{-7.14}}
& \small{\textcolor{BrickRed}{-20.64}}
& \small{\textcolor{BrickRed}{-25.68}}
& \small{\textcolor{BrickRed}{-6.64}}
\\

\midrule

GPT-5
& 29.77 & 38.52 & 62.62 & 68.51
& 45.24 & 56.58 & 78.90 & 77.41
\\

+ Joint multi-criterion
& 34.78 & 43.51 & 60.68 & 70.43
& 47.62 & 61.57 & 82.57 & 79.58
\\

\rowcolor{gray!6}
\small$\Delta$
& \small{\textcolor{ForestGreen}{+5.01}}
& \small{\textcolor{ForestGreen}{+4.99}}
& \small{\textcolor{BrickRed}{-1.94}}
& \small{\textcolor{ForestGreen}{+1.92}}
& \small{\textcolor{ForestGreen}{+2.38}}
& \small{\textcolor{ForestGreen}{+4.99}}
& \small{\textcolor{ForestGreen}{+3.67}}
& \small{\textcolor{ForestGreen}{+2.17}}
\\

\midrule

o4-mini
& 32.78 & 43.11 & 64.56 & 69.67
& 53.17 & 65.84 & 83.49 & 80.85
\\

+ Joint multi-criterion
& 36.79 & 40.12 & 58.74 & 68.82
& 45.24 & 55.87 & 77.06 & 75.61
\\

\rowcolor{gray!6}
\small$\Delta$
& \small{\textcolor{ForestGreen}{+4.01}}
& \small{\textcolor{BrickRed}{-2.99}}
& \small{\textcolor{BrickRed}{-5.82}}
& \small{\textcolor{BrickRed}{-0.85}}
& \small{\textcolor{BrickRed}{-7.93}}
& \small{\textcolor{BrickRed}{-9.97}}
& \small{\textcolor{BrickRed}{-6.43}}
& \small{\textcolor{BrickRed}{-5.24}}
\\

\midrule

Gemini-2.5-Pro
& 28.76 & 37.92 & 66.50 & 63.67
& 41.27 & 52.33 & 75.93 & 73.06
\\

+ Joint multi-criterion
& 32.11 & 41.12 & 62.62 & 67.06
& 35.71 & 40.21 & 56.88 & 70.96
\\

\rowcolor{gray!6}
\small$\Delta$
& \small{\textcolor{ForestGreen}{+3.35}}
& \small{\textcolor{ForestGreen}{+3.20}}
& \small{\textcolor{BrickRed}{-3.88}}
& \small{\textcolor{ForestGreen}{+3.39}}
& \small{\textcolor{BrickRed}{-5.56}}
& \small{\textcolor{BrickRed}{-12.12}}
& \small{\textcolor{BrickRed}{-19.05}}
& \small{\textcolor{BrickRed}{-2.10}}
\\

\bottomrule
\end{tabular}

\caption{Comparison of joint multi-criterion judgment against standard single-criterion judgment. Relative improvements are shown in  \textcolor{ForestGreen}{green}, decreases in \textcolor{BrickRed}{red}.}
\label{tab:appendix-experiment-joint-criterion}
\end{table*}
\renewcommand{\arraystretch}{1.0}

\definecolor{evalblue}{HTML}{EEF4FF}

\begin{table*}[!ht]
\begin{minipage}{1.0\textwidth}
\centering
\scalebox{0.94}{
\renewcommand{\arraystretch}{1.03}
\begin{tabular}{@{}p{2.5cm}@{}p{15cm}@{}}
\rowcolor{tbhead}
\toprule
\multicolumn{2}{l}{Case Study: Joint Multi-Criterion Failure of the \textit{o4-mini} Judge in Capturing Preference Conflicts} \\
\midrule
& \includegraphics[width=3.7cm]{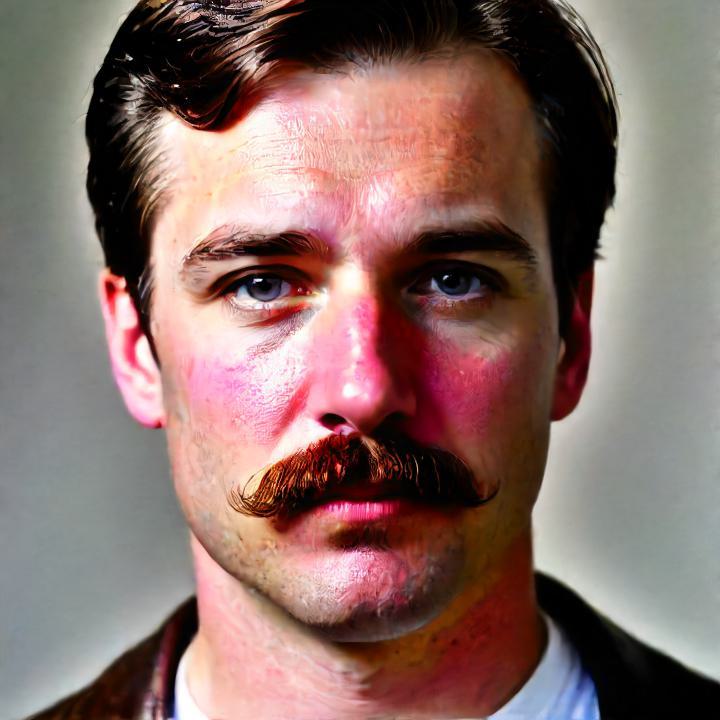} \\[0.3em]
Question &
Provide a single 10-year age bracket (e.g., “30s,” “40s,” etc.) and describe the man’s hair color and hairstyle based only on the image.
\\
\midrule
\arrayrulecolor{grid}

\rowcolor{tbshade}
\makecell[lt]{Response 1} &
Based on the image provided, the man appears to be in his \textcolor{critTextA}{30s}. His hair is dark brown and is \textcolor{critTextA}{styled in a classic fashion}, swept to the side, which complements his \textcolor{critTextC}{period attire}. The style is reminiscent of early 20th-century fashion, which aligns with the vintage feel of the image.
\\
\midrule
\makecell[lt]{Response 2 }
&
The man in the photo appears to be in his late 30s to early 40s. His hair is dark brown with a slight wave pattern. It is neatly styled, suggesting that he takes care of his appearance. The hair is cut short, and the parting on the side complements his overall look.
\\
\arrayrulecolor{critD}\midrule\arrayrulecolor{grid}
Ground Truth &
\vspace{-1.05em}
\begin{tabular}{p{5cm} p{3cm}}
\cellcolor{critA} Completeness and Coverage &  Resp. 1 $>$ Resp. 2 \\
\cellcolor{critC} Factuality / No Hallucination & Resp. 2 $>$  Resp. 1 \\
\end{tabular}
\\

\midrule
Single-Criterion \par Judgment \par (\textit{o4-mini}) &
\begin{minipage}[t]{\linewidth}
\vspace{-0.99em}
\fcolorbox{grid}{critA}{%
\parbox{\dimexpr\linewidth-2\fboxsep-2\fboxrule\relax}{%
\normalsize
\textbf{Completeness and Coverage}\par
The prompt asks for a single 10-year age bracket (e.g., “30s,” “40s,” etc.). Response 1 clearly states “in his 30s,” satisfying that requirement exactly. Response 2 instead gives “late 30s to early 40s,” which spans two decades and thus does not fulfill the prompt’s decade-long range specification.  - Both responses describe the hair color (“dark brown”) and hairstyle. However, Response 1 stays focused on the core question by succinctly noting the side-swept, classic style evident in the image. Response 2 adds extra inferences (“suggesting that he takes care of his appearance”) and splits the age range improperly, making it less precise with respect to the user’s request.  Because Response 1 fully and directly addresses both parts of the question—\textcolor{critTextA}{providing the correct decade-long age bracket plus hair color and style}—while Response 2 does not, Response 1 is the better answer under the completeness and coverage criterion. \par Response 1 is better. \boxed{\text{\textcolor{ForestGreen}{\cmark}}}
}}

\fcolorbox{grid}{critC}{%
\parbox{\dimexpr\linewidth-2\fboxsep-2\fboxrule\relax}{%
\normalsize
\textbf{Factuality / No Hallucination}\par
Response 1 introduces \textcolor{critTextC}{unsupported details about “period attire” and “early 20th-century fashion,”} neither of which are evident in the image—this is a visual hallucination. Response 2, by contrast, sticks to what is visible (age estimate spanning late 30s to early 40s, dark brown hair, slight wave, side part, neatly styled) without inventing extraneous context. 
Response 2 is better. \boxed{\text{\textcolor{ForestGreen}{\cmark}}}
}}
\end{minipage}
\\
\midrule
\rowcolor{evalblue}
Joint \par Multi-Criterion \par Judgment \par (\textit{o4-mini}) &
\normalsize
\textbf{Criterion: Completeness and Coverage} Response 1 directly answers both parts of the prompt within the requested format: it \textcolor{critTextA}{gives a single 10-year range (“30s”) and describes hair color (“dark brown”) and style (“classic fashion, swept to the side”), even tying in contextual cues (“period attire,” “vintage feel”)}. Response 2, by contrast, straddles two decades (“late 30s to early 40s”), which fails to provide a single 10-year band as requested, and although it describes hair color and neat styling, it does not anchor its observations as neatly to the prompt’s format. Judgment: Response 1 is better. \boxed{\text{\textcolor{ForestGreen}{\cmark}}}
\par \textbf{Criterion: Factuality / No Hallucination} Both responses correctly observe dark brown hair and a side part, but Response 2 introduces \textcolor{critTextA}{imprecision by spanning “late 30s to early 40s,” which contradicts the user’s instruction for a single decade}. Response 1 sticks to a plausible age (“30s”) based on the man’s appearance. Neither response hallucinates details about color, but Response 2’s age estimate violates the prompt’s constraints. Judgment: Response 1 is better. \boxed{\text{\textcolor{BrickRed}{\xmark}}}
\\
\bottomrule
\end{tabular}
\renewcommand{\arraystretch}{1.0}
}
\vspace{-1mm}
\captionof{table}{\textbf{In this case, \textit{o4-mini} correctly identifies the \textit{completeness vs.\ no-hallucination} conflict under single-criterion judgments but collapses to a uniform preference under joint multi-criterion judgment.} With single-criterion evaluation prompts, the model successfully adheres to each criterion independently and captures the preference disagreement, whereas under joint prompting, completeness dominates and the model fails to follow the no-hallucination criterion.}
\label{tab:openended_age_hair_conflict_batch_example}
\end{minipage}
\end{table*}

\end{document}